%% file: main.tex
\documentclass[10pt,journal,compsoc]{IEEEtran}
\usepackage{cite}
\ifCLASSINFOpdf
\else
\fi
\usepackage{makecell}
\usepackage{pifont}
\usepackage{subfiles}
\usepackage{graphicx}
\usepackage[colorlinks]{hyperref}
\usepackage{subfigure}
\usepackage[labelfont=bf]{caption}
\usepackage{amsfonts}
\usepackage{mathtools}
\usepackage{booktabs}
\usepackage{multirow}
\usepackage{nicematrix}
\usepackage{comment}
\usepackage{amssymb}
\newcommand{\Paragraph}[1]{\vspace{1mm} \noindent \textbf{#1.} \hspace{0mm}}
\usepackage{setspace}
\usepackage{algorithm}
\usepackage{algorithmic} 
\usepackage[ruled,vlined,algo2e]{algorithm2e}
\begin{document}

%\title{SkySight: Whole-Body Biometric Recognition in Aerial Surveillance}
%\title{HighSight: Whole-Body Biometric Recognition}
%\title{FarSight 2.0: End-to-End Physics-Driven Whole-Body Biometric System at Large Distance and Altitude}
\title{Person Recognition at Altitude and Range: Fusion of Face, Body Shape and Gait}
%-----------------------------------------------
\author{Feng~Liu,~\IEEEmembership{Senior Member,~IEEE,}
        Nicholas Chimitt,
        Lanqing~Guo,
        Jitesh~Jain,
        Aditya~Kane,
        Minchul~Kim, 
        Wes~Robbins,
        Yiyang~Su,
        Dingqiang~Ye,
        Xingguang~Zhang,
        Jie~Zhu,
        Siddharth~Satyakam,
        Christopher~Perry,
        Stanley~H.~Chan,~\IEEEmembership{Senior Member,~IEEE,}
        Arun~Ross,~\IEEEmembership{Senior Member,~IEEE,} Humphrey~Shi, Zhangyang~Wang, ~\IEEEmembership{Senior~Member,~IEEE}, Anil~Jain,~\IEEEmembership{Life~Fellow,~IEEE}, and Xiaoming~Liu,~\IEEEmembership{Fellow,~IEEE}
\IEEEcompsocitemizethanks{\IEEEcompsocthanksitem 
%Feng Liu is with the Department of Computer Science, Drexel University, Philadelphia, PA, 19104. \\
Feng Liu, Minchul Kim, Yiyang Su, Dingqiang Ye, Jie Zhu, Siddharth Satyakam, Christopher Perry, Arun Ross, Anil Jain and Xiaoming Liu are with the Department
of Computer Science and Engineering, Michigan State University, East Lansing, MI, 48824. \\
Nicholas Chimitt, Xingguang Zhang, Stanley H. Chan are with the School of Electrical and Computer Engineering, Purdue University, West Lafayette, IN, 47907. \\
Jitesh Jain, Aditya Kane, Humphrey Shi are with the School of Interactive Computing, Georgia Tech, Atlanta, GA, 30332. \\
Lanqing Guo, Wes Robbins, Zhangyang Wang are with the Department of Electrical and Computer Engineering, University of Texas at Austin, Austin, TX, 78712. \protect\\
E-mail: \{jain; liuxm\}@msu.edu
}

\thanks{Manuscript received April 19, 2005; revised August 26, 2015.}}

\markboth{Journal of \LaTeX\ Class Files,~Vol.~14, No.~8, August~2015}%
{Shell \MakeLowercase{\textit{et al.}}: Bare Demo of IEEEtran.cls for Computer Society Journals}

%% ---------------------------------------- %% 
\IEEEtitleabstractindextext{%
\begin{abstract}
We address the problem of whole-body person recognition in unconstrained environments. This problem arises in surveillance scenarios such as those in the IARPA Biometric Recognition and Identification at Altitude and Range (BRIAR) program, where biometric data is captured at long standoff distances, elevated viewing angles, and under adverse atmospheric conditions (\emph{e.g.}, turbulence and high wind velocity).
To this end, we propose \textbf{FarSight}, a unified end-to-end system for person recognition that integrates complementary biometric cues across face, gait, and body shape modalities. FarSight incorporates novel algorithms across four core modules: multi-subject detection and tracking, recognition-aware video restoration, modality-specific biometric feature encoding, and quality-guided multi-modal fusion. These components are designed to work cohesively under degraded image conditions, large pose and scale variations, and cross-domain gaps.
Extensive experiments on the BRIAR dataset, one of the most comprehensive benchmarks for long-range, multi-modal biometric recognition, demonstrate the effectiveness of FarSight. Compared to our preliminary system~\cite{liu2024farsight}, this system 
achieves a 34.1\% absolute gain in 1:1 verification accuracy (TAR@0.1\% FAR), a 17.8\% increase in closed-set identification (Rank-20), and a 34.3\% reduction in open-set identification errors (FNIR@1\% FPIR).
%, while delivering an end-to-end throughput of 5.0 FPS.
%achieves a 34.1\% improvement in verification (TAR@0.1\% FAR), a 17.8\% increase in closed-set identification (Rank-20), and a 34.3\% gain in open-set false negatives (FNIR@1\% FPIR), relative to our preliminary system~\cite{liu2024farsight}. 
%
Furthermore, FarSight was evaluated in the 2025 NIST RTE Face in Video Evaluation (FIVE), which conducts standardized face recognition testing on the BRIAR dataset. These results establish FarSight as a state-of-the-art solution for operational biometric recognition in challenging real-world conditions.

\end{abstract}

}

%~\footnote{no. of subjects in probe and gallery, acquisition device}

\maketitle
\IEEEdisplaynontitleabstractindextext
\IEEEpeerreviewmaketitle

%%% ---------------------- ------------------- ----------------- %%

\subfile{sec/sec_1_intro.tex}

\subfile{sec/sec_2_prior.tex}

%%
\subfile{sec/sec_3_method.tex}

%%

\subfile{sec/sec_4_exp.tex}
%%

\subfile{sec/sec_5_conclusion.tex}
%%

%%% ---------------------- ------------------- ----------------- %%

\ifCLASSOPTIONcaptionsoff
  \newpage
\fi

{\small
\bibliographystyle{IEEEtran}
\bibliography{refs}
}
\subfile{sec/sec_6_bio.tex}

\end{document}

%% file: sec/sec_1_intro.tex
\IEEEraisesectionheading{\section{Introduction}\label{sec:intro}}
\IEEEPARstart{U}{nconstrained} biometric recognition at long distances and elevated viewpoints is crucial for a variety of applications, including law enforcement, border security, wide-area surveillance, and public media analytics~\cite{gong2011person,zheng2016person,jain2024speaker}. 
Among existing approaches, whole-body biometric recognition~\cite{liu2024farsight,CoNAN,WholeDIR,bolme2024data,BRIAR,BRIAR2} has become a central focus in this domain, as it captures a rich combination of anatomical and behavioral traits—such as facial appearance, gait and body shape—offering greater resilience to occlusion, degradation, and modality loss than single-modality systems. 
Despite its potential, deploying whole-body recognition systems in real-world scenarios remains technically demanding. High-performing systems must not only incorporate robust multi-modal biometric modeling, but also support modules for precise person detection and tracking, enhancement of low-quality imagery, mitigation of atmospheric turbulence, and adaptive fusion strategies to handle unreliable data. %Meeting these requirements is essential for achieving reliable recognition performance in unconstrained environments.

%--------------------- Figure 1 --------------------%
\begin{figure*}
\centering
\includegraphics[width=1.0\linewidth]{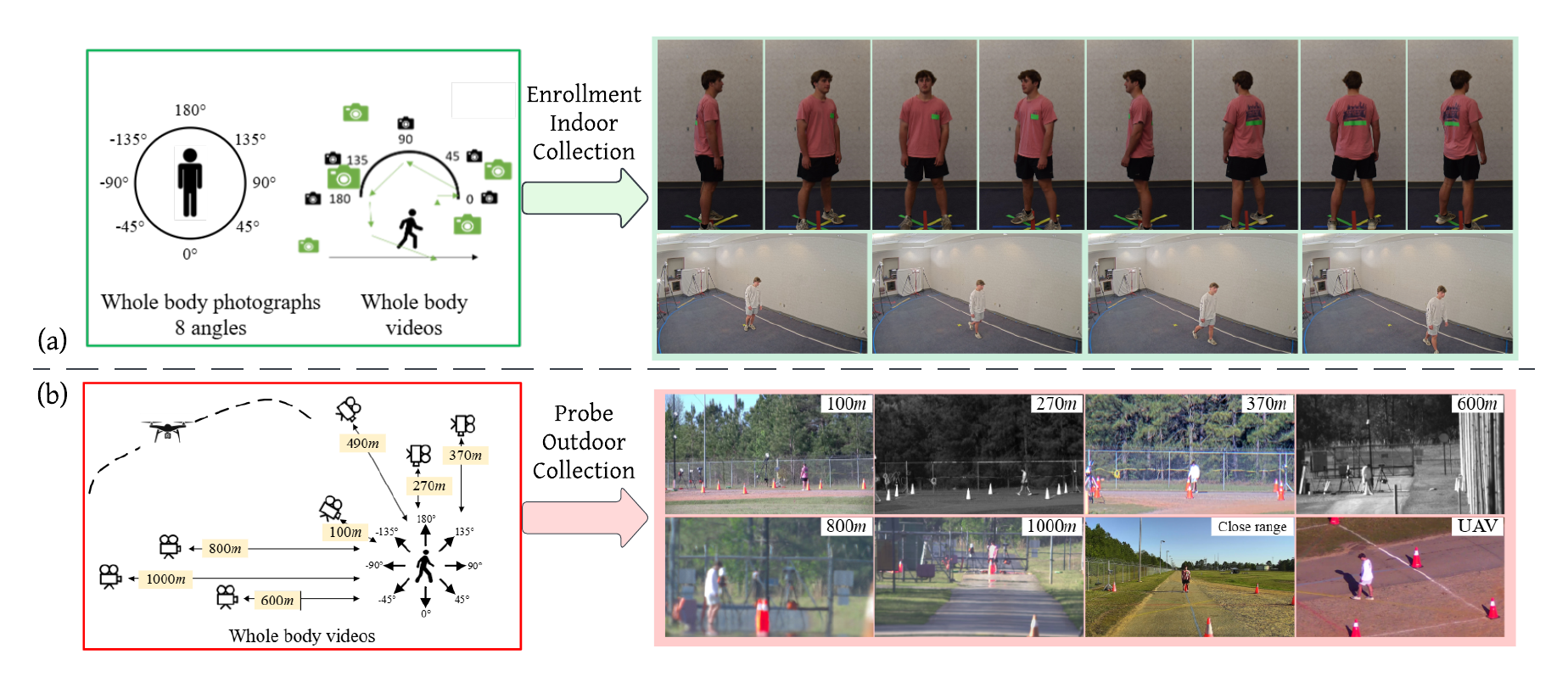}
\caption{Illustration of the IARPA BRIAR whole-body image capture scenarios. (a) Enrollment Indoor Collection: High-quality still images and videos captured from multiple viewpoints under controlled conditions. (b) Probe Outdoor Collection: Videos captured in outdoor environments at varying distances and elevation angles, with challenging factors such as atmospheric turbulence. These settings reflect the real-world conditions encountered in long-range biometric recognition. \emph{Permission granted by the subject for use of imagery in publications}.}
\label{fig:briar_dataset}
\end{figure*}
%------------------------------------------------%

%%
To develop and evaluate biometric systems that meet these demands, it is essential to have access to datasets that reflect the full complexity of real-world surveillance conditions.
%Advancing the development of whole-body biometric recognition requires not only robust algorithms but also a large and comprehensive dataset tailored for long-range and high-altitude tasks. \gt{what is the reference or support for this statement?}
The IARPA Biometric Recognition and Identification at Altitude and Range (BRIAR) program\footnote{\url{https://www.iarpa.gov/research-programs/briar}} is a collective effort~\cite{BRIAR,BRIAR2} in this direction, fostering the development of biometric systems capable of performing reliably in these unconstrained scenarios. 
Fig.~\ref{fig:briar_dataset} illustrates the BRIAR whole-body image capture scenarios, comprising controlled indoor enrollment collections and challenging outdoor probe collections.
These scenarios simulate the real-world challenges in person recognition, including: 
\textbf{\emph{(i)}} low-quality video frames caused by long-range capture (up to 1000 meters) and atmospheric turbulence, with refractive index structure constant ranging from $C_n^2 = 10^{-17}$ to $10^{-14} \, \text{m}^{-2/3}$; \textbf{\emph{(ii)}} large yaw and pitch angles (up to $50^\circ$) from elevated platforms (drones) at altitudes up to 400 meters; \textbf{\emph{(iii)}} degraded feature sets due to low visual quality, where the Inter-Pupillary Distance (IPD) ranges between 15--100 pixels; \textbf{\emph{(iv)}} the complexity of open-set search, where probe images must be matched against galleries containing distractors;  and \textbf{\emph{(v)}} a significant domain gap caused by limited training data and the diversity of real-world conditions.

To address the challenges posed by unconstrained, long-range biometric recognition, we propose \textbf{FarSight}, an integrated end-to-end system designed for robust person recognition using multi-modal biometric cues.
FarSight combines face, gait, and body shape modalities to ensure recognition performance even when individual cues are unreliable or degraded. The system comprises four tightly coupled modules, each addressing a critical component of the recognition pipeline:
\textbf{\emph{(1)}} A \textbf{multi-subject detection and tracking module} that accurately localizes individuals in video sequences captured under dynamic, cluttered, and low-resolution conditions. 
\textbf{\emph{(2)}} A \textbf{recognition-aware video restoration module} that mitigates visual degradation—particularly due to turbulence and long-range blur—by jointly optimizing image quality and biometric fidelity. 
\textbf{\emph{(3)}} A \textbf{biometric feature encoding module} that extracts robust representations for each modality, leveraging recent advances in large vision models and modality-specific architectural designs.
\textbf{\emph{(4)}} A \textbf{quality-guided multi-modal fusion module} that adaptively integrates scores across modalities, accounting for variable input quality and partial observations.

A preliminary version of our system~\cite{liu2024farsight} was previously presented at the IEEE/CVF Winter Conference on Applications of Computer Vision (WACV 2024). Building on that foundation, we have substantially upgraded each module to improve recognition performance across verification, closed-set identification, and open-set search tasks. The current system also incorporates key architectural enhancements to support lower latency, reduced memory usage, and improved scalability.
%under the evolving requirements of the BRIAR program.
Below, we summarize the major improvements introduced in each module of the updated FarSight system:

\vspace{0mm}
\;\textbullet\ \textbf{Multi-Subject Detection and Tracking:} 
Our initial system~\cite{liu2024farsight} employed a joint body-face detector based on R-CNN~\cite{girshick2014rcnn,Wan_2021_ICCV}, which lacked support for multi-subject tracking and exhibited high inference latency. To address these limitations, we introduce two key upgrades:
First, we adopt a dual-detector framework using BPJDet~\cite{bpjdet} for coarse body-face localization followed by verification via YOLOv8~\cite{yolov8_ultralytics} to reduce false positives. This replacement improves both detection accuracy and runtime efficiency.
Second, we develop PSR-ByteTrack, an enhanced multi-subject tracker built on ByteTrack~\cite{zhang2022bytetrack}. PSR-ByteTrack mitigates issues such as ID switches, fragmented tracklets, and reidentification failures by introducing a patch-based retrieval mechanism that maintains subject-specific appearance features in memory. %This allows for accurate ID reassignment based on visual similarity, even when bounding-box associations fail due to occlusion or re-entry.

%\gtc{FarSight~\cite{liu2024farsight} leveraged a R-CNN~\cite{girshick2014rcnn} based joint body-face detector~\cite{wan2021body} to detect humans for a given video. Although accurate for single-subject settings, it did not support tracking multiple subjects, making it impractical for real-world scenarios. Overcoming the mentioned limitation, we present a more accurate and robust detection-tracking module in FarSight 2.0. Specifically, we propose two key advancements. Firstly, we introduce a \textit{YOLO~\cite{yolov8_ultralytics} based dual-detector system} with initial body-face detections from a YOLOv5~\cite{yolov5_ultralytics} based BPJDet~\cite{bpjdet} followed by verification with a YOLOv8~\cite{yolov8_ultralytics}  to reduce false-positives from BPJDet. Moreover, using YOLO-based detectors also reduces the latency of our module compared to the R-CNN-based detector from FarSight~\cite{liu2024farsight}. Secondly, we equip our module with a \textit{multi-subject tracker}. We introduce PSR-ByteTrack, a novel enhancement on top of ByteTrack~\cite{zhang2022bytetrack} to tackle issues such as ID-switch, tracklet fragmentation, and disappearing \& reappearing subjects, which arise frequently in multi-subject settings. Specifically, we develop a general patch-feature similarity-based postprocessing technique to accurately update incorrect track-id assignments from ByteTrack, owing to its bounding-box association-guided subject tracking approach. We populate a patch-memory with patch features for each subject to accurately update track-ids based on the subject's appearance.}

%%
\vspace{0mm}
\;\textbullet\ \textbf{Recognition-Aware Video Restoration}:
We introduce the Gated Recurrent Turbulence Mitigation (GRTM) network, a novel video-based restoration model tailored for long-range, turbulence-degraded imagery. A lightweight classifier is used to trigger restoration selectively, reducing unnecessary computation and avoiding potential feature distortion.
A key contribution of this system is its tightly coupled restoration-recognition co-optimization framework which
%Traditional restoration methods prioritize visual quality, often introducing artifacts that distort biometric features and degrade recognition accuracy. In contrast, our approach 
integrates recognition objectives directly into the restoration training process, guiding the model to enhance features critical for identity discrimination. %This task-aware optimization substantially improves recognition performance over standalone restoration methods while minimizing identity hallucination, highlighting the importance of aligning enhancement strategies with downstream biometric tasks.

%
% \vspace{1mm} \item \textbf{\textcolor{blue}{Recognition-aware} video restoration}: We develop the \textcolor{blue}{Gated Recurrent Turbulence Mitigation network (GRTM), a novel video-based turbulence mitigation model.} We apply restoration only when necessary by training a
% video classifier to automatically determine whether an input
% video should undergo the restoration process.
% %Finally, we co-optimize our physics-based video restoration model with a joint restoration-recognition objective so that the video restoration model improves biometric performance without hallucinating facial features.
% \textcolor{blue}{A unique contribution of FarSight 2.0 is the introduction of a tightly coupled restoration-recognition co-optimization framework. Traditional image restoration approaches typically focus solely on visual quality metrics, often inadvertently introducing hallucinated or distorted biometric features, thus impairing downstream recognition accuracy. By contrast, our approach explicitly integrates biometric recognition objectives directly into the restoration training process, ensuring restoration efforts specifically enhance features most beneficial for biometric matching. Our experiments demonstrate that this joint optimization strategy significantly improves recognition performance compared to standalone restoration methods, while effectively minimizing identity hallucination. Such strategic integration not only fortifies system accuracy under realistic atmospheric distortions but also underscores the necessity for task-aware restoration in biometric systems.}

%%
\vspace{0mm}
\;\textbullet\ \textbf{Biometric (Face, Gait and Body Shape) Feature Encoding:}
We upgrade each modality-specific model with task-aligned architectural improvements and training strategies tailored to the challenges of long-range, unconstrained biometric recognition.
\emph{i) Face}: 
We propose KP-RPE~\cite{kim2024keypoint}, a keypoint-dependent relative position encoding technique 
which
%. This method enhances the ViT backbone's ability to manage unseen affine transformations, 
significantly improves the handling of misaligned and low-quality facial images.
\emph{ii) Gait}: 
We introduce BigGait~\cite{ye2024biggait}, the first gait recognition framework based on large vision models (LVMs). This method shifts from task-specific priors to general-purpose visual knowledge, improving gait recognition across diverse conditions.
\emph{iii) Body shape}: We propose CLIP3DReID~\cite{liu2024distilling}, which significantly enhances body matching capabilities by synergistically integrating linguistic descriptions with visual perception. This method leverages the pre-trained CLIP model to develop discriminative body representations, effectively improving recognition accuracy.

\vspace{0mm}
\;\textbullet\ \textbf{Quality-Guided Multi-Modal Fusion:} 
We propose Quality Estimator (QE), a general approach for assessing modality quality 
%using pseudo quality loss without the need for human-labeled quality data, and a multi-modal biometric recognition framework with 
and a learnable score-fusion method guided by modality-specific quality weights called Quality-guided Mixture of score-fusion Experts (QME) to enhance score-fusion performance.

\vspace{0mm}
\;\textbullet\ \textbf{Open-Set Search:} 
%To address the challenges of real-world open-set biometric recognition—where probe identities may not exist in the gallery—
We introduce a new training strategy~\cite{su2024open} that explicitly incorporates non-mated subjects. This approach aligns the training objective with open-set conditions, enabling the model to distinguish between enrolled and unknown identities. As a result, it significantly improves open-set recognition accuracy while also enhancing closed-set performance through better generalization.

\vspace{0mm}
\;\textbullet\ \textbf{System Integration:} 
We incorporate several system-level enhancements which
%to align with the updated BRIAR API specifications for testing. Key upgrades 
include: \emph{i)} automated multi-GPU containerization, enabling each GPU to process client requests independently; and \emph{ii)} support for multi-subject probe videos, allowing a single input to produce multiple subject track entries. %Additional improvements include full codebase refactoring, GPU memory and affinity optimizations, and raster export tools for failure analysis and data preprocessing

%The pipeline architecture is determined in part by the API specified by Oak Ridge National Laboratories. Subsequent to the publication of our preliminary system, this API was upgraded to meet BRIAR program requirements for phase II and III testing and evaluation. We implement these changes in the FarSight system, along with various other performance improvements. There are two primary API updates that are implemented in this release.  The first is the implementation of multi-processing in the containers such that a separate container instance is launched automatically for each GPU to process requests from the client independently, eliminating the need to launch multiple container instances manually. The second is the support for multi-subject probe videos.  A single probe video can now generate multiple track entries correlating to multiple subjects in the video input.  Additional improvements include a full codebase refactoring, support for 2-mean-2 scoring fusion, GPU memory optimizations, improved GPU affinity, and intermediate raster export to assist in failure case analysis and training data pre-processing.

%%
In summary, our contributions of the proposed \textbf{FarSight} system include:

$\diamond$ Utilizing a dual YOLO-based detection approach, coupled with our PSR-ByteTrack for robust, accurate, and low-latency multi-subject detection and tracking. 

%$\diamond$ Explicitly modeling the physics of imaging through turbulence and image degradation and integrating physics-based models into deep learning for image/video restoration. 

%$\diamond$ Co-optimizing our physics-based video restoration model with a joint restoration-recognition objective so that the video restoration model improves biometric performance without hallucinating facial features.

$\diamond$ A physics-informed video restoration module (GRTM) that explicitly models atmospheric turbulence and integrates a task-driven, recognition-aware optimization framework to enhance identity-preserving image quality.

$\diamond$ Effective feature encoding for face, gait, and body shape, augmented by a large vision model framework. This approach integrates a novel approach to open-set search and multimodal feature fusion, significantly enhancing recognition performance across diverse scenarios.
%$\diamond$ Modality-specific biometric encoders for face, gait, and body shape, upgraded with architectural advances and LVM-guided supervision to improve robustness under misalignment, resolution loss, and domain shift.

%$\diamond$ A quality-guided multi-modal fusion framework (QME) that dynamically weights modality contributions using pseudo-quality estimation, improving recognition performance under degraded or missing inputs.

%$\diamond$ An open-set training strategy that explicitly incorporates non-mated identities, aligning model optimization with real-world deployment objectives and improving both open-set and closed-set performance.

$\diamond$ Scalable system integration with automated per-GPU multi-processing and support for multi-subject probe handling, in accordance with updates to the API specification.

$\diamond$ Comprehensive evaluation on the BRIAR dataset (protocol v5.0.1) and independent validation through the 2025 NIST RTE Face in Video Evaluation (FIVE)~\cite{nistfrte2025}, confirming FarSight's state-of-the-art performance in operational biometric recognition under real-world conditions.

%% file: sec/sec_2_prior.tex
\section{Related Work}\label{sec:prior}

%\Paragraph{Whole-Body Biometrics Recognition \textcolor{red}{Yiyang}}
\Paragraph{Whole-Body Person Recognition}
Whole-body person recognition integrates multiple biometric traits, such as face, gait, and body shape, to achieve state-of-the-art identification accuracy in challenging scenarios. 
This holistic approach contrasts sharply with traditional biometric systems that typically focus on a single modality~\cite{arcface,kim2022adaface,fan2023opengait,CAL,connor2018biometric,zhang2020learning,liu2018disentangling,improving-face-recognition-from-hard-samples-via-distribution-distillation-loss,fan-feature-adaptation-network-for-surveillance-face-recognition-and-normalization}. By integrating multiple modalities, FarSight overcomes the limitations of individual traits while harnessing their complementary strengths. 
%Unlike traditional biometric systems that focus primarily on a single trai this holistic approach mitigates the weaknesses of individual traits and takes advantage of the complementary strengths of each trait. 
For instance, while face recognition can struggle with severe pose changes and poor lighting, gait analysis can be affected by variations in walking speed and clothing. Similarly, body shape provides consistent cues, but can be altered by variations in clothing and pose.
%CoNAN,bolme2024data,WholeDIR
Recent studies~\cite{liu2024farsight,CoNAN,WholeDIR} have increasingly adopted holistic systems that integrate detection, image restoration, and biometric analysis.
%have increasingly adopted holistic systems with detection, restoration, and biometrics. 
However, many existing systems still rely on relatively small-scale networks trained on restricted datasets and fail to fully capitalize on the potential synergies among different biometric modalities and system components.
This motivates the development of an integrated system that jointly optimize across the entire recognition pipeline. Our work builds on this trend by incorporating large vision models, task-aware restoration, open-set training, and adaptive multi-modal fusion into a scalable, end-to-end system evaluated under real-world environment.

%\Paragraph{Physics Modeling of Imaging through Turbulence \textcolor{red}{Purdue}}
\Paragraph{Physics Modeling of Imaging through Turbulence}
Atmospheric turbulence is a major source of image degradation in long-range and high-altitude person identification, significantly impairing both visual clarity and biometric recognition accuracy. This challenge necessitates realistic simulation methods to support both the training to yield robust recognition systems and the development of effective restoration algorithms.
%Turbulence is a significant factor in long-range and high-altitude person identification, affecting image clarity and recognition accuracy. This degradation necessitates robust atmospheric turbulence simulation techniques not only to enhance the training of biometric systems but also for effective image restoration under such conditions. 
%
Simulation techniques span a wide spectrum—from physics-based models grounded in computational optics~\cite{Hardie_2017_a}, which provide high fidelity at the cost of computational expense, to computer vision-based methods~\cite{Milanfar_2013_a} that prioritize efficiency but often lack physical grounding. Intermediate approaches include brightness function-based simulations~\cite{Vorontsov_2005_a} and learning-based techniques~\cite{Miller_2021_a}, though the latter differs from runtime constraints, particularly in deep learning settings~\cite{Mao_2021_a}.
%Atmospheric turbulence simulation spans from computational optics \cite{Hardie_2017_a, Roggemann_2012_a, Roggemann_1995_a, Schmidt_2010_a} that rely on expensive wave computations to computer vision-oriented approaches \cite{Milanfar_2013_a, Lau_2019_a, Chak_2021_a} that offer speed, but arguably lack physical foundations. Other approaches fall between these two extremes, such as simulations based on brightness functions \cite{Vorontsov_2005_a, Lachinova_2007_a, Lachinova_2017_a} or learning-based alternatives \cite{Miller_2019_a, Miller_2021_a}, although speed remains a bottleneck for deep learning approaches~\cite{Mao_2021_a}. 
%
To balance realism and efficiency, we adopt a turbulence model based on random phase distortions represented by Zernike polynomials. Our approach synthesizes turbulence effects by applying numerically derived convolution kernels to a clean image and injecting white noise, producing a realistic degraded observation. %This method offers a practical trade-off between physical accuracy and computational efficiency, enabling large-scale training and evaluation of turbulence-aware biometric systems.
%We model atmospheric turbulence using a random field of phase distortion represented by Zernike coefficients. This model employs numerically derived convolution kernels to simulate the effect of turbulence on a ground truth image, taking into account the distortions and adding white noise to replicate the turbulence-degraded image. This method bridges the gap between high computational fidelity and the need for operational speed in practical applications.

\begin{figure*}
\centering
\includegraphics[width=1.0\linewidth]{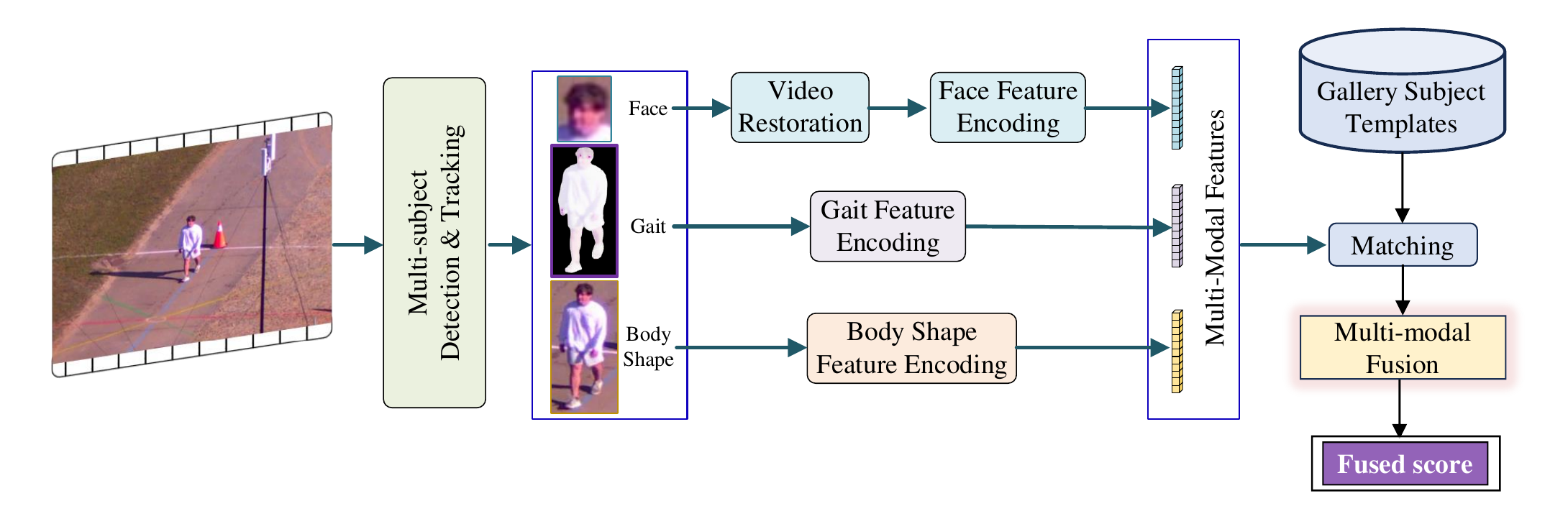}
%\caption{The proposed FarSight system incorporates six modules: \emph{detection and tracking, image restoration, face, gait, and body shape encoding, and multi-modal biometric fusion}.}
%\vspace{-3mm}
\caption{Overview of the proposed \textbf{FarSight} system, which comprises four modules: (i) multi-subject detection and tracking, (ii) recognition-aware image restoration, (iii) modality-specific encoding for face, gait, and body shape, and (iv) quality-guided multi-modal biometric fusion.}
\label{fig:overview}
\end{figure*}

\Paragraph{Image Restoration for Biometric Recognition}
Biometric recognition relies on extracting robust features from diverse visual inputs. When image quality is suboptimal, restoration techniques can enhance image fidelity and, in turn, improve recognition performance. However, such methods may inadvertently alter identity by hallucinating features or degrade accuracy by introducing artifacts. 
Additionally, conventional
restoration pipelines often optimize for perceptual metrics
such as PSNR or SSIM, which poorly reflect recognition
accuracy~\cite{jain1997practicing, liu2018image, vidalmata2020bridging, yang2020advancing}.
Under atmospheric turbulence, reconstruction has been found beneficial~\cite{jaiswal2023physics}. While these efforts predominantly rely on single-frame data, whereas multi-frame turbulence mitigation can lead to more stable and reliable restoration \cite{Zhang_2022_a, Zhang_2024_datum}. %Notably, generative blind face restoration methods~\cite{Nair_2023_a} can lead to identity shift issues~\cite{Wang_2021_GFPGAN}. 
%
%Additionally, conventional restoration pipelines often optimize for perceptual metrics such as PSNR or SSIM, which poorly reflect recognition accuracy~\cite{jain1997practicing, wang2016studying, liu2018image, vidalmata2020bridging, yang2020advancing}.
%
In contrast, FarSight introduces a deterministic multi-frame restoration framework co-optimized with biometric recongition accuracy objectives. This strategy explicitly aligns restoration with recognition accuracy, preserving identity features while mitigating the risk of visual hallucination.

% To address these concerns, the FarSight system adopts a deter
% \textcolor{blue}{Prior works addressing image restoration in biometric systems have predominantly utilized single-stage restoration methods guided by generic quality metrics such as PSNR or SSIM. However, these metrics lack explicit correlation with biometric performance, often resulting in restorations visually pleasing but suboptimal for identity recognition tasks \cite{jain1997practicing,wang2016studying,liu2018image,vidalmata2020bridging,yang2020advancing}. Furthermore, standalone restoration modules may inadvertently hallucinate biometric features, severely impairing recognition reliability, especially at long-range surveillance scenarios with significant atmospheric turbulence. In contrast, our co-optimization framework, integral to FarSight 2.0, directly ties restoration effectiveness to biometric recognition metrics, such as TAR and FNIR. By explicitly optimizing restoration networks using a hybrid loss function combining perceptual quality measures and biometric-specific AdaFace classification loss, we achieve enhanced restoration that is task-driven rather than purely aesthetic-driven. Our comprehensive experiments validate this approach, demonstrating concrete biometric performance improvements over conventional restoration methods—specifically, a measurable gain of around 3.8\% in verification accuracy (TAR@0.1\% FAR) and clear reductions in false negatives during open-set identification scenarios.}

\Paragraph{Person Detection and Tracking}
%\textbf{Detection.} 
% Human body and face detection and association is critical in developing person recognition in video~\cite{liu2024farsight, arcface}. 
Detecting and associating persons across multiple frames is critical for developing accurate person recognition systems. 
Early approaches \cite{Chi_Zhang_Xing_Lei_Li_Zou_2020, zhang2019double} use an R-CNN-based detector with multiple heads for independent body and face detection, followed by a matching module. BFJDet \cite{Wan_2021_ICCV} proposes a framework for converting any one- or two-stage detector to support body and face detection. More recently, PairDETR \cite{Ali_2024_CVPR} uses a DETR-inspired bipartite framework to match body and face bounding boxes. FarSight~\cite{liu2024farsight} uses a Faster R-CNN~\cite{ren2015faster} to jointly detect human bodies and faces. Due to recent advances in real-time detection algorithms, particularly the YOLO series~\cite{yolov5_ultralytics, li2022yolov6, yolov8_ultralytics, yolox2021}, BPJDet develops a joint detection algorithm using YOLOv5~\cite{yolov5_ultralytics} and an association decoding to match body with face. Farsight leverages BPJDet as the main detector and uses YOLOv8~\cite{yolov8_ultralytics} to eliminate false body detections.
%

%\noindent
%\textbf{Tracking.} 
Tracking by association (of bounding boxes or segmentation masks)~\cite{zhang2022bytetrack, unitrack, pujara2022deepsort, chen2018real} is an established practice for multi-object tracking.   
%Multi-object tracking has grown into its own field of study, and newer approaches augment the classical approaches. For instance, \cite{du2023strongsort} introduces an appearance-free link model to improve upon the classical DeepSORT \cite{pujara2022deepsort} approach. On the other hand, \cite{lv2024diffmot} incorporates a real-time diffusion model to account for complex trajectories. ByteTrack~\cite{zhang2022bytetrack} introduces the notion of caching low-confidence detection boxes and reviving them as required. Our tracker builds on top of ByteTrack and implements a patch-based post-processing technique for accurate multi-object tracking that benefits downstream biometric modules.
Under the association paradigm, ByteTrack~\cite{zhang2022bytetrack} caches low-confidence bounding boxes, resulting in an accurate tracker for both high and low-confidence detections. Owing to its impressive performance for multi-subject tracking, we use ByteTrack as our base tracker equipped with an appearance-aware patch-based post-processing technique for accurate track-id assignment, leading to robust person recognition.

%\Paragraph{Multi-Modal Biometric Fusion \textcolor{red}{Jie}}
\Paragraph{Multi-Modal Biometric Fusion}
Score-level fusion is a widely used approach in multi-modal biometric systems, where similarity scores from individual modalities—such as face, gait, or body shape—are combined to form a final person recognition decision. Traditional techniques include normalization-based methods (\emph{e.g.}, Z-score, Min-Max) followed by mean, max, or min score fusion~\cite{jain2005score}. Likelihood ratio-based methods~\cite{nandakumar2007likelihood} have also been proposed to provide probabilistic interpretability.
Despite their simplicity, these fusion methods often fail to account for modality-specific reliability or dynamic quality variations in the input. A key challenge lies in determining optimal modality alignment and weighting under real-world intra-person variations. Some recent works have moved toward feature-level fusion~\cite{aung2022multimodal}, combining information across modalities (\emph{e.g.}, face and gait) to exploit cross-modal correlations. However, these approaches may suffer from representation incompatibility or lack robustness to missing modalities.
To address these limitations, our approach introduces a quality-guided score-fusion framework that dynamically weighs each modality’s contribution based on estimated quality of the probe. %Specifically, we combine pseudo-quality estimators with a learnable mixture-of-experts (QME) model, enabling adaptive and generalizable fusion across varying surveillance conditions.

%Score-level fusion combines similarity scores generated by individual modalities to form a final decision. Each modality yields its own similarity score, which indicates the likelihood that two inputs (e.g., images or videos) correspond to the same individual. Common score fusion methods include the Z score and Min-Max normalization.~\cite{nandakumar2007likelihood} introduces likelihood ratio-based score fusion. Ross \textit{et al.} propose a fusion of mean, max, or min score, where the final score is determined by averaging or the highest or lowest score among multiple matches~\cite{jain2005score}. However, challenges remain in determining the optimal alignment and weighting for each modality and identifying the most effective fusion strategy. We have explored a better way to assess the contribution of each modality and developed a more generalizable score-fusion method. To achieve this, we utilize quality weights for input modalities along with multiple experts to guide the score-fusion process. Some works focus on multi-modal feature fusion like~\cite{aung2022multimodal} by combining information from both face and gait information. 

%\Paragraph{Open-set Search \textcolor{red}{Yiyang}}
\Paragraph{Open-Set Biometric Search}
Open-set search is a critical requirement in whole-body biometric systems, where a probe must be matched to an enrolled subject, if present, or rejected if not enrolled in the gallery. Despite its practical importance, most prior work in whole-body biometrics has focused on closed-set recognition, with limited attention to explicitly modeling open-set dynamics.
%Open-set search is critical in whole-body biometrics as it requires identifying a probe as one of the subjects enrolled in the gallery while rejecting those not enrolled. Despite this, most whole-body biometric models have mainly focused on general tasks, such as designing loss functions or specialized architectures, without addressing open-set challenges directly. 
A common baseline is the Extreme Value Machine (EVM)\cite{gunther2017toward}, which estimates the likelihood that a probe belongs to each gallery subject and rejects low-confidence matches. In our work\cite{su2024open}, we introduced a training strategy that explicitly simulates open-set conditions by incorporating non-mated identities during training. This alignment between training and evaluation improves generalization and boosts performance in both open-set and closed-set scenarios.

%Two notable methods related to open-set search include EVM~\cite{gunther2017toward}, which uses extreme value machines to estimate the probability of a probe belonging to each gallery subject, and rejects probes with low probabilities. The approach in \cite{su2024open}, improves open-set performance by simulating open-set scenarios during training to align it with testing conditions. This alignment enhances performance in both open-set and closed-set scenarios.

%% file: sec/sec_3_method.tex
\section{Proposed Method}\label{sec:method}

\subsection{Overview of the FarSight System}
As illustrated in Fig.~\ref{fig:overview}, the proposed FarSight system consists of four tightly integrated modules: multi-subject detection and tracking, recognition-aware image restoration, modality-specific feature encoding (face, gait, and body shape), and a quality-guided multi-modal fusion module. These components are orchestrated within a unified, end-to-end framework designed to address the real-world challenges outlined in Sec.~\ref{sec:intro}—namely, long-range capture, pose variation, degraded imagery, and domain shift. 

The system is optimized for scalability and efficiency, handling galleries of approximately 99,000 still images and 12,000 video tracks, while maintaining an end-to-end processing speed of 7.0 FPS on 1080p videos using an NVIDIA RTX A6000 GPU.
%scalability and operational efficiency. 
It supports dynamic batch sizing for GPU resource management and communicates with external systems via an API built on Google RPC. Video inputs are specified through configuration files, and extracted biometric features are exported in HDF5 format for downstream evaluation and scoring. The recognition pipeline begins with person detection and tracking. For each tracklet, cropped frames are passed to the gait and body shape encoders. Simultaneously, facial regions undergo restoration to mitigate degradation before entering the face encoder. Each probe consists of a single video segment, while gallery enrollments—comprised of multiple videos and stills—are aggregated into a single feature vector per modality.

\subsection{Multi-Subject Detection and Tracking}

\subsubsection{Person Detection}
To enable reliable subject localization under unconstrained settings, we adopt a dual-detector strategy that combines BPJDet~\cite{bpjdet} and YOLOv8~\cite{yolov8_ultralytics} for robust body-face detection. BPJDet serves as the primary detector, independently predicting body and face bounding boxes and associating them by computing the inner IoU—defined as the intersection over the face bounding box area—between candidate body-face pairs.
%\gtc{We introduce a dual detector approach for obtaining accurate body-face detections for every person in a given probe. We use BPJDet proposed by Zhou \emph{et al.}~\cite{bpjdet} as our primary body-face detector. During inference, BPJDet first individually predicts the body and face bounding boxes and associates them by calculating the inner IoU (the denominator is the area of the face box instead of the body box) between different pairs of bodies and faces and pairing those with the highest inner IoU.}

During development, we observe that BPJDet occasionally produces false positives in the presence of distractor objects (\emph{e.g.}, traffic cones or robotic fixtures), which negatively impact downstream biometric encoding. To mitigate this, we introduce a verification step using YOLOv8~\cite{yolov8_ultralytics}. Specifically, a detection from BPJDet is retained only if YOLOv8 also detects a corresponding body within a confidence threshold of 0.7. This cross-verification step significantly reduces false positives without compromising recall. Following body-face detection, subjects are temporally associated across frames using our PSR-ByteTrack tracker, described below.
%\gtc{During the development phase, we find that BPJDet tends to output false-positive body detections for probes containing distractor objects like traffic cones or robot cameras which are harmful to the biometric modules' performance. To tackle the false positives, we verify the BPJDet's body detections with those from a more accurate YOLOv8~\cite{yolov8_ultralytics}, ensuring minimal false positive detections. Specifically, we only retain a BPJDet detection if it is also returned from the YOLOv8 model with a certain confidence threshold (0.7).}

%\gtc{Once we obtain the body-face detections, we use our PSR-ByteTrack method to associate the detected subjects across multiple frames in a probe.}

\noindent\textbf{Throughput Optimization.} While accurate, the naive integration of BPJDet and YOLOv8 introduced computational bottlenecks due to redundant preprocessing. Both detectors share similar input transformations, leading to redundant CPU operations and suboptimal GPU utilization. To address this, we implemented two key optimizations:
(i) a unified preprocessing pipeline to eliminate shared steps across detectors; and
(ii) a GPU-efficient pipeline, which reduces CPU load. These improvements yield a 5× increase in throughput on a single GPU without impacting detection accuracy.

%\gtc{\textbf{Throughput Optimization.} Although our dual detector pipeline is accurate, we observe that a naive integration of the two models results in sub-optimal throughput. After intensive analysis, we found that both the detectors share most preprocessing steps. This not only results in redundant operations in the pipeline, but also causes massive CPU bottlenecks. Consequently, we make a couple of optimizations: (i) firstly, we deduplicate pre-processing operations between the two models and develop a unified pre-processing pipeline; and (ii) secondly, we develop a GPU-friendly pre-processing pipeline, resulting in $5\times$ higher throughput on a single GPU without any drop in performance.} 

\subsubsection{Person Tracking}
For multi-subject tracking, we build upon the ByteTrack algorithm~\cite{zhang2022bytetrack}, which uses a two-stage association mechanism—first linking high-confidence detections, followed by low-confidence ones. While ByteTrack performs well under general conditions, we observed two key limitations in long-range surveillance settings: (i) frequent ID switches during occlusions, and (ii) fragmented tracklets when reidentifying subjects who temporarily exit and reenter the scene. To address these issues, we introduce Patch Similarity Retrieval ByteTrack (PSR-ByteTrack), a patch-based post-processing framework that refines ByteTrack’s output using appearance-based reidentification. 

\begin{figure}[t]
    \centering
    \includegraphics[width=\linewidth]{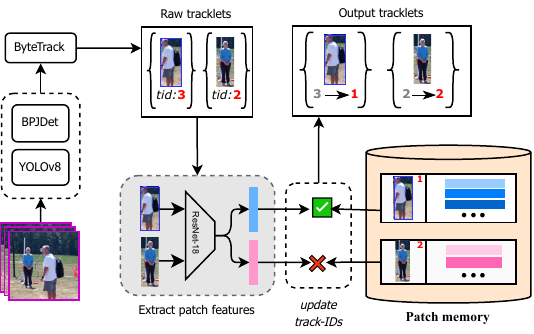}
    % \vspace{-0.8cm}
    %\caption{\gtc{We introduce a patch-based post-processing to correct the tracklets generated by ByteTrack~\cite{zhang2022bytetrack} with patch-similarity driven retrieval (PSR) and update of track-IDs. We first encode the patches in the proposed tracklet using a ResNet-18 model. We then retrieve the most similar patches from the patch memory and update the respective track IDs if the initial track IDs do not match with their retrieved counterparts. This way, we keep track of the subjects' appearance and correct the tracklets generated by ByteTrack.}}
    \caption{Overview of the multi-subject detection and tracking in FarSight. A dual-detector approach combines BPJDet\cite{bpjdet}  for body-face localization and YOLOv8\cite{yolov8_ultralytics} for false positive suppression. Detected subjects are then associated across frames using PSR-ByteTrack~\cite{zhang2022bytetrack}, which refines ByteTrack outputs through patch similarity-based retrieval and track ID correction. This ensures consistent tracking under occlusions, subject re-entry, and long-range degradation.}
    % \caption{Our Patch Similarity-based Retrieval ByteTrack system refines tracklets generated by ByteTrack~\cite{zhang2022bytetrack}. We utilize YOLOv8 alongside BPJDet to eliminate false positives and employ a patch memory system that updates track IDs based on patch similarity and bounding box intersection over union (IoU).}
    \label{fig:psr}
\end{figure}

As illustrated in Fig.~\ref{fig:psr}, we maintain a patch memory, where each entry corresponds to a track ID and contains ResNet-18\cite{resnet}-encoded features from body patches.
The pipeline proceeds as follows: (i) Initial tracklets are obtained from ByteTrack using body detections. (ii) For each new detection, if the associated track ID does not yet exist in the memory, we store its patch feature. (iii) At every $N$ frames, new patches are appended to account for temporal appearance changes.
(iv) For each incoming patch, we compute the mean squared error (MSE) with stored features in the memory and assign the track ID with the lowest error, provided the similarity exceeds a pre-defined threshold. (v) Detections with low similarity to all existing entries are treated as new subjects and assigned new IDs.

%\subsection{Video Restoration and Recognition Co-Training}
\subsection{Recognition-Aware Video Restoration}
\label{sec:cotraining}

\subsubsection{Atmospheric Turbulence Modeling and Simulation}
\label{sec:simulation}
Image degradation from atmospheric turbulence presents a critical challenge in long-range face recognition, introducing spatial and temporal varying blur. The severity of this distortion is influenced by propagation distance, camera parameters, and turbulence strength~\cite{Roggemann_1996_a, chan2023computational}. To train models that are robust under such conditions, we synthesize degradation-free image pairs using Zernike polynomial-based turbulence simulation\cite{Chimitt_2020_a, Mao_2021_a, Chimitt_2022_a}, applied to both static\cite{zhou2017places} and dynamic~\cite{Safdarnejad_2015_a, Jin_2021_a} scenes. Our simulations span a range of turbulence strengths (\emph{e.g.}, $D/r_0 \in [1, 10]$) and camera configurations (\emph{e.g.}, f-number, sensor size), providing diverse training data aligned with FarSight’s real-world acquisitions.
%Image degradation by atmospheric turbulence manifests as a spatio-temporally varying blur, influenced by various factors such as camera parameters, propagation distance, and turbulence strength \cite{Roggemann_1996_a, chan2023computational}. To synthesize data, we utilize Zernike-based simulation \cite{Chimitt_2020_a, Mao_2021_a, Chimitt_2022_a} to simulate accurate ground truth and degraded image pairs using dynamic \cite{Safdarnejad_2015_a, Jin_2021_a} static scenes \cite{zhou2017places} across a wide range of turbulence strengths, \emph{e.g.}, $D/r_0 \in [1, 10]$, and camera parameters such as $f$-number, sensor size, etc. This ensures that generated turbulence sequences are sufficiently diverse to match the various use cases of our system.

\subsubsection{GRTM Network and Selective Restoration}
To enhance facial imagery under severe atmospheric distortions, we designed an efficient Gated Recurrent Turbulence Mitigation (GRTM) Network based on the state-of-the-art video turbulence mitigation framework DATUM \cite{Zhang_2024_datum}. To improve efficiency and robustness, we removed the optical flow alignment in \cite{Zhang_2024_datum} since it takes significant computational resources and may introduce artifacts to harm downstream recognition tasks. To further reduce the potential negative impact caused by restoration artifacts, we employ a video classifier trained on real-world videos and their restored pairs to indicate whether or not the restoration could potentially improve recognition performance. 

\subsubsection{Co-Optimization of Restoration and Recognition}
\begin{figure}[t]
    \centering
    \includegraphics[width=0.99\linewidth]{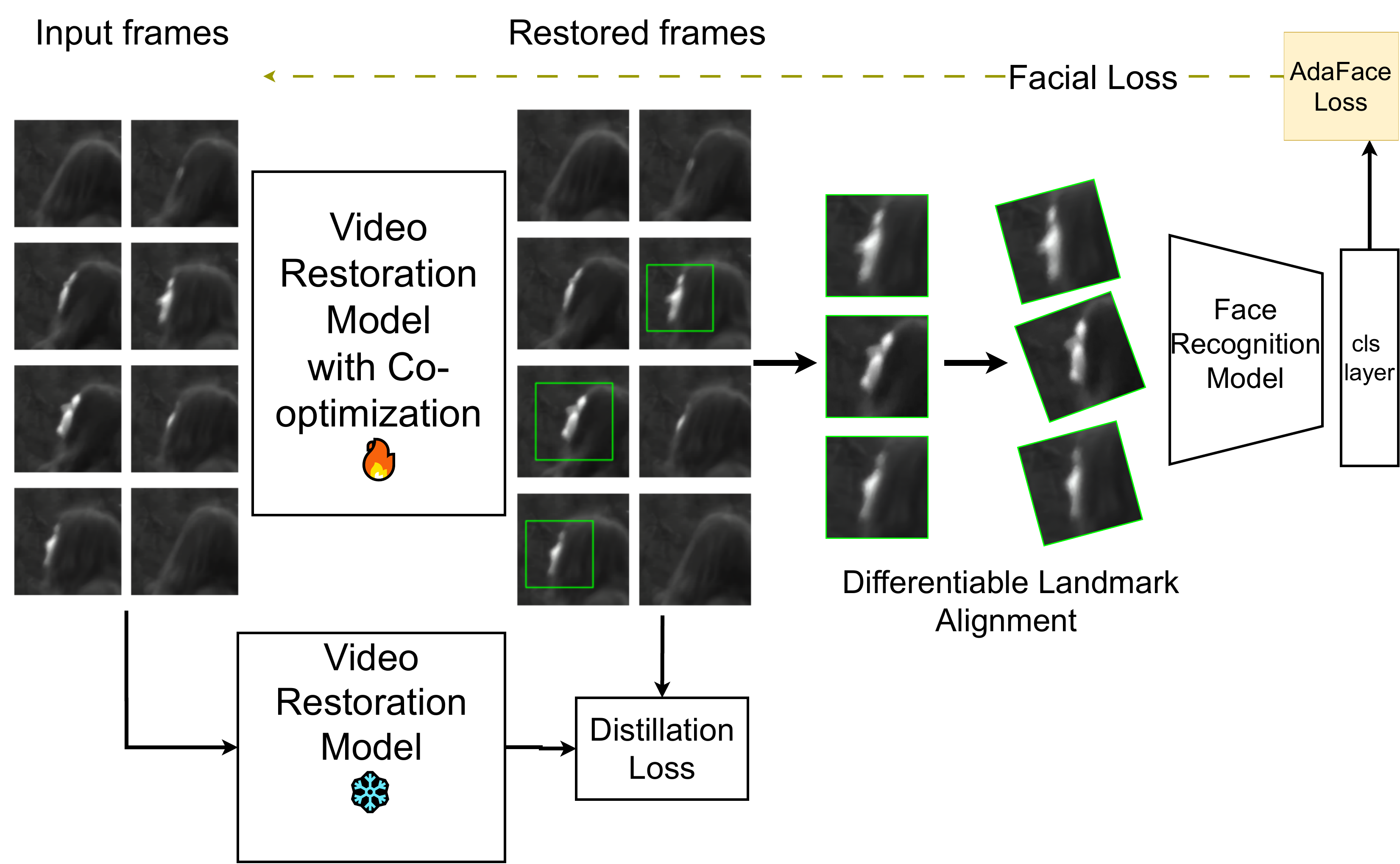}
    \caption{Training pipeline for the proposed restoration-recognition co-optimization framework. A distillation loss between siamese-twin models and our face recognition model helps us define a loss for the face recognition model. As shown, not all frames may have detections and only frames with detections are used in $\mathcal{L}_{\text{adaface}}$.}
    \label{fig:co_optimization}
\end{figure}

Conventional restoration models typically optimize generic visual metrics (\emph{e.g.}, PSNR, SSIM), which do not align with biometric recognition goals and may hallucinate identity-altering features. To overcome this, 
%Traditional image restoration methods, although visually effective, typically optimize generic quality metrics such as PSNR or SSIM and often inadvertently introduce hallucinated or distorted biometric features. To overcome this limitation and to explicitly optimize the interplay between restoration and recognition,
we propose a restoration-recognition co-optimization framework, illustrated in Fig.~\ref{fig:co_optimization}. The framework adopts a teacher-student configuration, where a frozen teacher model provides high-quality visual references, and the student model is fine-tuned to jointly optimize for both visual fidelity and identity preservation.
%we introduce a restoration-recognition co-training strategy, illustrated in Fig.~\ref{fig:co_optimization}. Our co-training framework employs a teacher-student model configuration, where a pretrained and stable teacher model remains fixed throughout training to provide high-quality restoration references. The student model, initialized from the teacher, is subsequently fine-tuned using a combination of restoration quality and face recognition metrics, enabling the restoration network to directly adapt to the nuanced demands of the recognition task.

Formally, the combined optimization objective for this co-training process is defined as follows:
\begin{equation}
\mathcal{L}_{\text{Co-op}} = \lambda \mathcal{L}_{\text{distill}} + \mathcal{L}_{\text{adaface}},
\end{equation}
where $\mathcal{L}_{\text{distill}}$ is the distillation loss that preserves the original restoration ability by minimizing the distance between the outputs of the teacher and student restoration models, effectively preserving the visual quality and realism of the restored images. Concurrently, $\mathcal{L}_{\text{adaface}}$ \cite{kim2022adaface} introduces a biometric-specific face classification loss to the co-training process. This component explicitly guides the restoration model toward enhancing facial features that contribute directly to improved identity discrimination capabilities.

The proposed joint optimization strategy enables each restored and aligned frame to be evaluated with respect to both visual quality and identity preservation. Through iterative feedback, the restoration model learns to prioritize visual features that are critical for accurate biometric recognition, while suppressing details that may introduce ambiguity or identity drift. In contrast to conventional methods that emphasize perceptual appeal, our approach ensures that restorations are not only visually coherent but also optimized to enhance recognition performance.
%The proposed joint optimization strategy uniquely allows each restored and aligned frame to be assessed and refined with respect to both image quality and identity preservation. Through iterative feedback, the restoration model dynamically learns which visual details are essential for accurate biometric recognition, and which may be safely de-emphasized or avoided entirely to prevent identity confusion. Unlike conventional methods, our approach thus ensures that the enhancements applied by the restoration model are not merely visually appealing but explicitly beneficial for biometric accuracy.

%Empirical results further validate our approach, demonstrating substantial improvements in biometric recognition accuracy compared to standalone restoration methods. Specifically, our co-trained model achieves an average improvement of approximately 3.8\% in TAR at 0.1\% FAR and significantly reduces false negative identification rates (FNIR) at 1\% FPIR. Qualitative analysis also confirms that our method successfully minimizes the introduction of artificial biometric features, a common shortcoming of standard restoration-only methods. Consequently, our co-training approach directly bridges the gap between visual restoration effectiveness and biometric recognition performance, providing a robust and practically advantageous solution for challenging real-world surveillance scenarios.

\subsection{Enhanced Biometric Feature Encoding with Large Vision Models}

%\subsubsection{Face \textcolor{red}{Minchul}}
\subsubsection{Face}
Conventional face recognition models 
%State-of-the-art models 
often struggle to extract meaningful facial features, particularly due to their reliance on properly aligned face images. 
To address this limitation, we incorporate the Keypoint Relative Position Encoding (KP-RPE)~\cite{kim2024keypoint} mechanism, which directly manipulates the attention mechanism in the Vision Transformer (ViT) model. By encoding relative positions of facial keypoints, KP-RPE enhances the model's robustness to misalignment and unseen geometric affine transformations. 
%This innovative approach ensures better feature extraction and improves performance across diverse and challenging datasets.

\textbf{Relative Position Encoding (RPE).}  
Relative Position Encoding (RPE), first introduced in~\cite{shaw2018self} and later refined in~\cite{dai2019transformer,huang-etal-2020-improve}, encodes sequence-relative position information to enhance self-attention mechanisms. Unlike absolute position encoding, RPE considers the relative spatial relationships between input elements, making it particularly useful for vision and language tasks. The modified self-attention mechanism incorporates relative positional embeddings $\mathbf{R}^Q_{ij}$, $\mathbf{R}^K_{ij}$, and $\mathbf{R}^V_{ij}$ into query-key interactions, where each $\mathbf{R}_{ij}$ is a learnable vector that encodes the relative distance between the $i$-th query and the $j$-th key or value. These embeddings allow attention scores to be adjusted based on sequence relative distances rather than fixed positions. Various distance metrics, such as the quantized Euclidean distance. %and separate $x,y$ distances, 
have been explored to compute these relationships~\cite{ramachandran2019stand, wu2021rethinking}. 

\begin{figure}[t]
    \centering
    \includegraphics[width=0.85\linewidth]{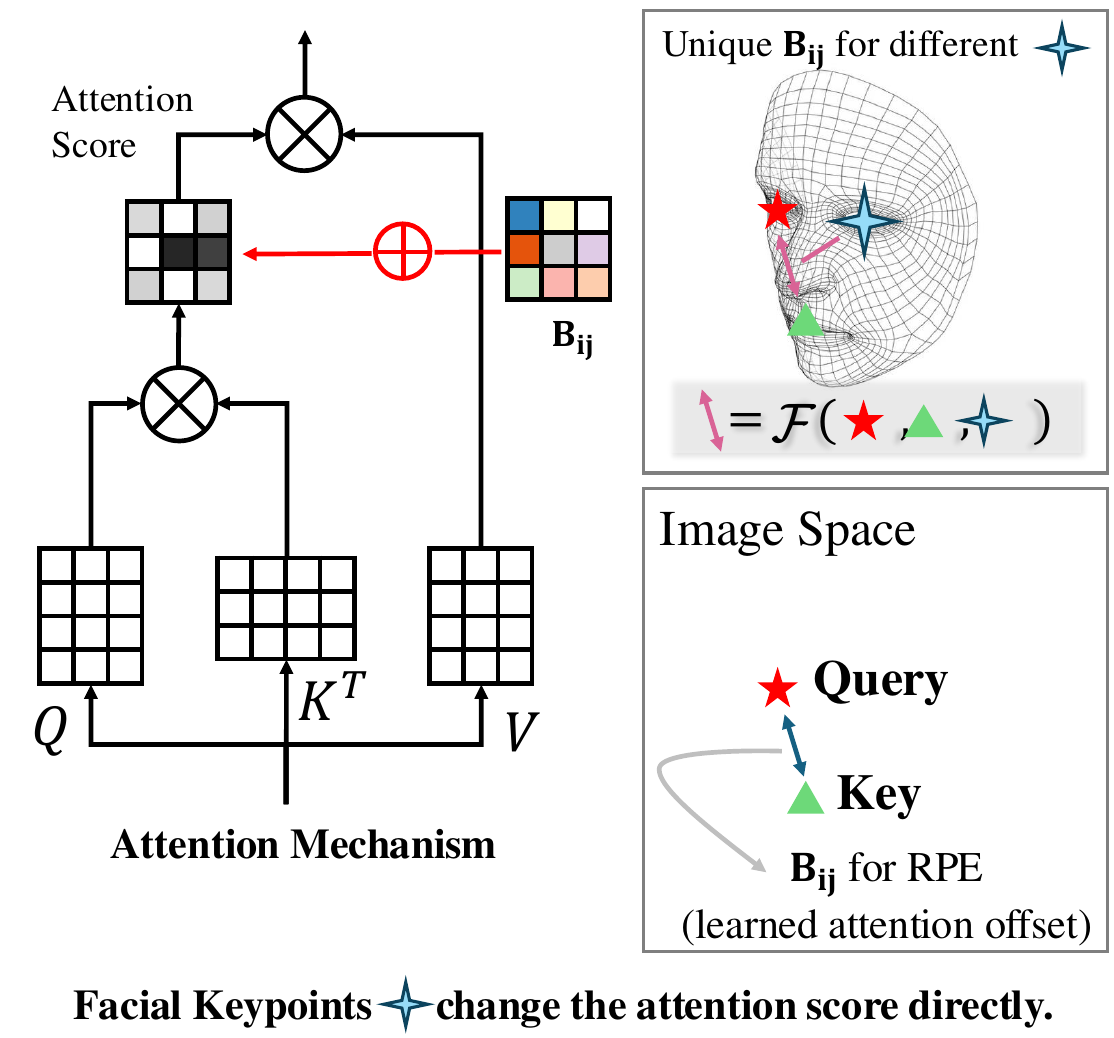}
    \caption{Illustration of keypoint relative position encoding (KP-RPE)~\cite{kim2024keypoint}. In standard RPE, the attention offset bias is computed based on the distance between the query $Q$ and the key $K$. In KP-RPE, the RPE mechanism is further enhanced by incorporating facial keypoint locations, allowing the RPE to dynamically adjust to the orientation and alignment of the image. }
    \label{fig:fig-kprpe}
\end{figure}

\textbf{Keypoint Relative Position Encoding (KP-RPE)}  
KP-RPE modifies the conventional RPE by incorporating keypoint information into the positional bias matrix $\mathbf{B}_{ij}$. Instead of making the distance function $d(i, j)$ explicitly dependent on keypoints, which limits efficiency due to pre-computability constraints, the matrix $\mathbf{B}_{ij}$ is defined as a function of keypoints: $\mathbf{B}_{ij} = \mathcal{F}(\mathbf{P})[d(i, j)]$. The function $\mathcal{F}(\mathbf{P})$ transforms keypoints into a learnable offset table, ensuring that the attention mechanism adapts based on keypoint-relative relationships. The final formulation enhances standard RPE by allowing the offset function to be relative to both the query-key positions and keypoints. This allows the RPE to be dependent on the image contents' position, making the model robust to misalignment. In Fig.~\ref{fig:fig-kprpe}, we provide an illustration of KP-RPE. 

%--------------------- Figure 1 --------------------%
%--------------------- Figure 1 --------------------%

% \noindent\textcolor{blue}{KP-RPE}

%\subsubsection{Gait \textcolor{red}{Dingqiang}}
\subsubsection{Gait}
\begin{figure}[t]
\centering
\includegraphics[width=1.0\linewidth]{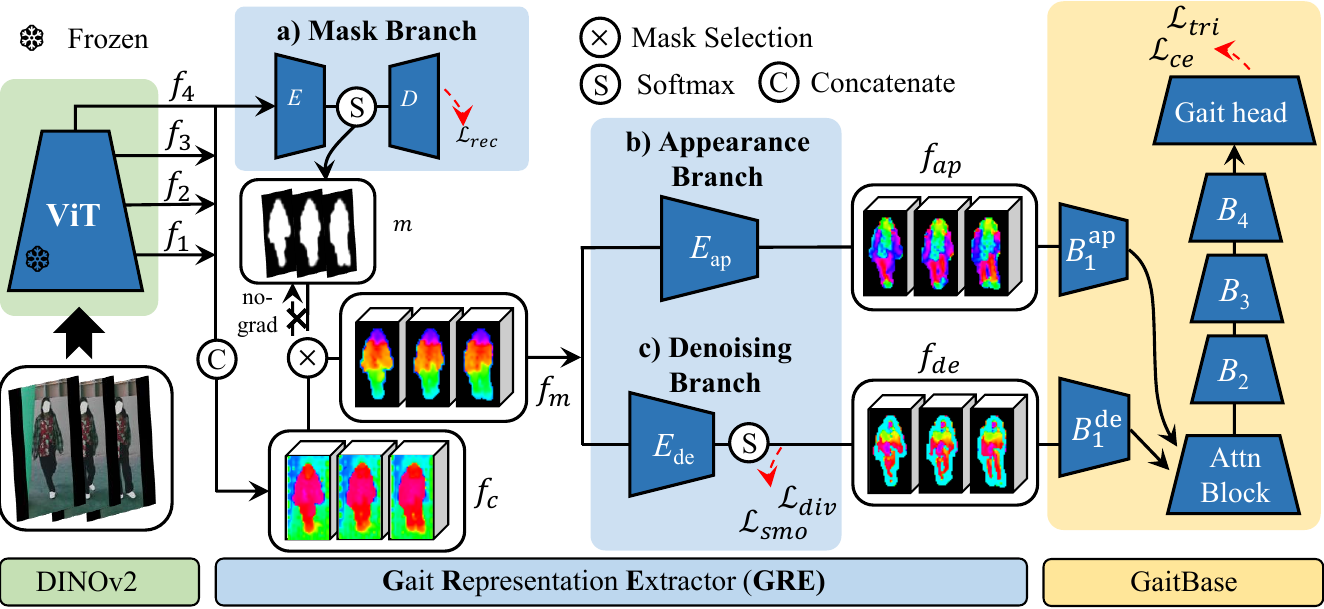}
\caption{Workflow of \textbf{BigGait}~\cite{ye2024biggait}. We adopt DINOv2\cite{oquab2023dinov2} as the upstream model to generate the feature maps: $f_1$, $f_2$, $f_3$, $f_4$ by various stages of the ViT backbone with the corresponding semantic hierarchy spanning from low to high levels. The gait representation extractor (GRE) comprises three branches for background removal, feature transformation, and denoising. An improved GaitBase is used for gait metric learning. }
\label{fig:biggait_overview}
\end{figure}

Conventional gait recognition methods predominantly rely on multiple upstream models driven by supervised learning to extract explicit gait features, such as silhouettes and skeleton points.
Breaking away from this trend, we introduce the BigGait~\cite{ye2024biggait} method, which leverages all-purpose knowledge generated by powerful Large Vision Models (LVMs) to replace traditional gait representations. 
As illustrated in Fig.~\ref{fig:biggait_overview}, we design three branches to extract gait-related representations from LVMs in an unsupervised manner.
This cutting-edge gait method achieves state-of-the-art performance in both within-domain and cross-domain evaluation.

BigGait processes all frames of an input RGB video in parallel.
To maintain accurate body proportions, it applies a Pad-and-Resize technique, resizing each detected body region to $448\times224$ pixels before feeding it into the upstream model.
The upstream DINOv2~\cite{oquab2023dinov2} is a scalable ViT backbone, selecting ViT-S/14 (21M) and ViT-L/14 (302M) for BigGait-S and BigGait-L.
The resized RGB image is split into $14 \times 14$ patches, which yields tokenized vectors of dimension $32 \times 16$. As shown in Fig.~\ref{fig:biggait_overview}, $f_1$, $f_2$, $f_3$ and $f_4$ are feature maps generated by various stages of the ViT backbone with the corresponding semantic hierarchy spanning from low to high levels. We concatenate these four feature maps along the channel dimension to form $f_c$.
%
%The tokens of various layers are sampled uniformly and spatially concatenated to form feature maps $f_4$ and $f_c$.
%
Formally, the feature maps $f_4$ and $f_c$ are processed through the Mask, Appearance, and Denoising branches.
%Mask Branch, the Appearance Branch, and the Denoising Branch.
\\
\noindent \textbf{Mask Branch.}
This branch acts as an auto-encoder that generates a foreground mask to suppress background noise using $f_4$: 
\begin{equation}
    \begin{aligned}
    m &= \text{softmax}(E(f_4)) \\ 
    \bar{f}_4 &= D(m) \\ 
    L_{rec} &= \left\| f_4 - \bar{f}_4 \right\|_2 ,
    \end{aligned}
\vspace{-2.mm}
\end{equation}
where $E$ and $D$ denote linear convolution layers with a $1\times1$ kernel and output channel with dimensionality of $2$ and $384$, respectively.
The foreground mask $m$ is then used to mask out background regions in $f_c$, yielding a foreground segmentation feature $f_m$:
\begin{equation}
    \begin{aligned}
    f_{m} = m \cdot f_{c} ,
    \end{aligned}
\end{equation}
where ``$\cdot$" denotes the multiplication operator.
\\
\noindent \textbf{Appearance Branch.}
This branch extracts the body shape characteristics from $f_m$:
\begin{equation}
    f_{ap} = E_{\text{ap}}(f_m) ,
\end{equation}
with $E_{\text{ap}}$ being a linear convolution layer with a $1\times1$ kernel and an output channel dimension of $C$.
\\
\noindent \textbf{Denoising Branch.}
To suppress high-frequency texture noise and obtain a skeleton-like gait feature, this branch employs both a smoothness loss $L_{smo}$ and a diversity loss $L_{div}$.
Specifically, the smoothness loss is:
\begin{equation}
    \begin{aligned}
        f_{de} &= \text{softmax}(E_{\text{de}}(f_m)) \\
        \mathcal{L}_{smo} &= | \text{sobel}_\text{x} * f_{de} | + | \text{sobel}_\text{y} * f_{de}|  ,
    \end{aligned}
    \label{equ:smo}
\end{equation}
where $E_{\text{de}}$ comprises a non-linear block formed by a $1\times1$ convolution, batch normalization, GELU activation, followed by an additional $1\times1$ convolution.
The diversity loss is:
\begin{equation}
    \begin{aligned}
        p_i &= \text{sum}(f_{de}^i) / \sum_{i=1}^{C} \text{sum}(f_{de}^i) \\
        \mathcal{L}_{div} &= \text{log}C + \sum_{i=1}^{C} p_i\text{log}p_i ,
    \end{aligned}
\end{equation}
%where $f_{de}^i$ and $p_i$, respectively, represent the activation map of the $i$-th channel and the corresponding frequency to entire activations. 
where $f_{de}^i$ represents the activation map of the $i$-th channel and $p_i$ is the proportion of activation for the $i$-th channel relative to the total activation across all channels.
The constant term $(\text{log}C)$ denotes the maximum entropy and is included to prevent negative loss. 
Finally, we fuse $f_{ap}$ and $f_{de}$ using attention weights:
\vspace{-2mm}
\begin{equation}
    \begin{aligned}
        f_{fusion} = Attn(\text{B}_1^{ap}(f_{ap}), \text{B}_1^{de}(f_{de})),
    \end{aligned}
\vspace{-2.mm}
\end{equation}
where $Attn$ is an attention block, following~\cite{fan2024skeletongait}, and the $f_{fusion}$ will be fed into GaitBase~\cite{fan2023opengait}.

%--------------------- Figure 1 --------------------%
%--------------------- Figure 1 --------------------%

\subsubsection{Body Shape}
\begin{figure}[t]
\centering
\includegraphics[width=1.0\linewidth]{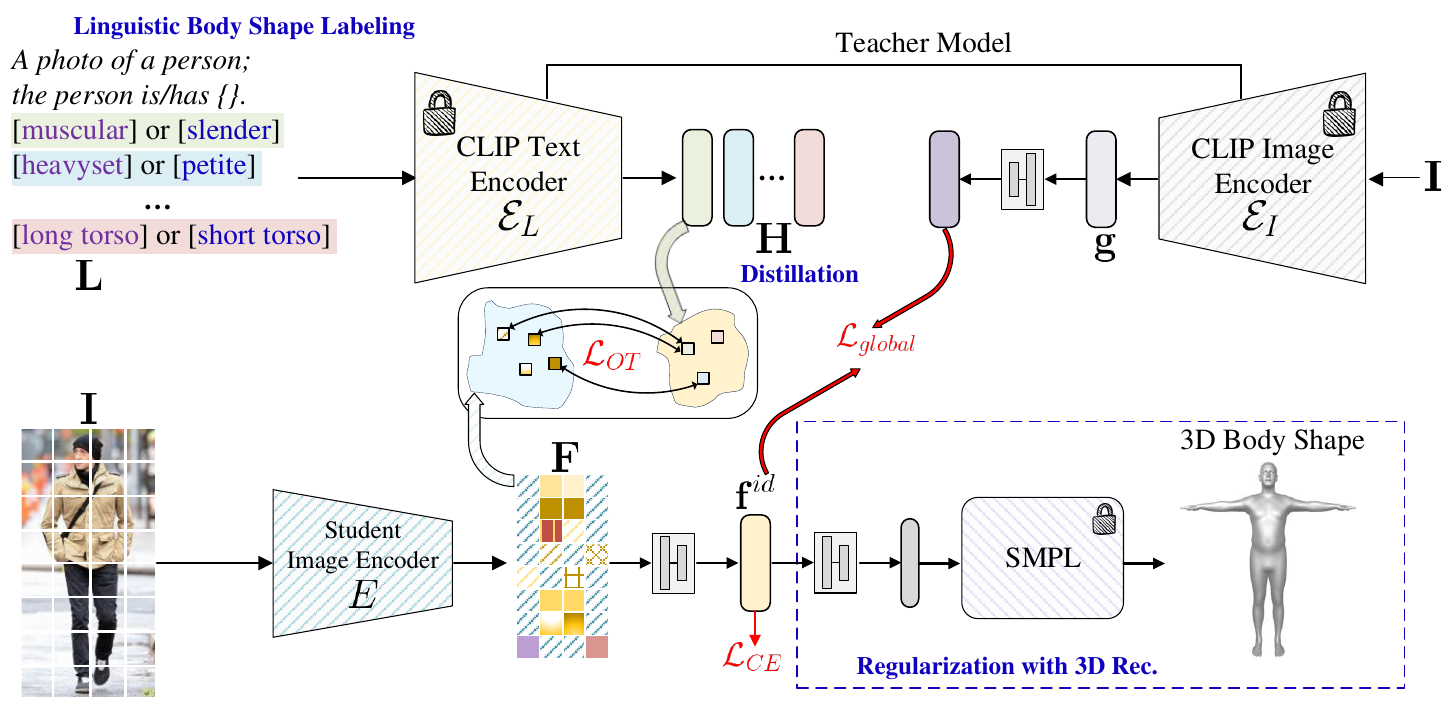}
\caption{Overview of the proposed CLIP3DReID~\cite{liu2024distilling} consisting of CLIP-based linguistic body shape labeling, dual distillation from CLIP, and regularization with 3D reconstruction. Incorporating these three modules into the person ReID framework enables us to learn discriminative body shape features.}
\label{fig:body_overview}
\end{figure}

To overcome the limitations of appearance-based attributes, such as clothing and color, we introduce CLIP3DReID~\cite{liu2024distilling}, a novel approach that significantly enhances the encoding of body shape features. As illustrated in Fig.~\ref{fig:body_overview}, this method leverages the pretrained CLIP model for knowledge distillation, integrating linguistic descriptions with visual perception for robust person identification. CLIP3DReID automatically labels body shapes with linguistic descriptors, employs optimal transport to align local visual features with shape-aware tokens from CLIP's linguistic output, and synchronizes global visual features with those from the CLIP image encoder and the 3D SMPL identity space. This integration achieves state-of-the-art results in person ReID.

%--------------------- Figure 1 --------------------%
%------------------------------------------------%

%%
Formally, for each mini-batch of $B$ training samples, denoted as $\{(\mathbf{I}_i, y_i, \mathbf{L}_{i})\}_{i=1}^{B}$, the input consists of human images $\mathbf{I}_i$,
the identity label of the image $y_{i}$, and a set of linguistic descriptors of body shape $\mathbf{L}_{i}$. We denote the \textbf{pre-trained} and \textbf{frozen} CLIP teacher text and image encoders as $\mathcal{E}_{L}$ and $\mathcal{E}_{I}$, respectively. The focus of our optimization is the student's visual encoder, represented as $E$. 

The CLIP teacher image encoder $\mathcal{E}_{I}$ processes the input image $\mathbf{I}$ and generates a feature vector $\mathbf{g}\in\mathbb{R}^{d}$. In the language component, the CLIP teacher text encoder $\mathcal{E}_{L}$, working with a set of $M$ linguistic body shape descriptors $\mathbf{L}=\{l_m\}^{M}_{m=1}$, outputs text feature sets $\mathbf{H}=\{\mathbf{h}_{m}\}_{m=1}^{M}\in\mathbb{R}^{M\times d}$.  %, where $[..]$ denotes the concatenation.
The student image encoder $E$ also takes $\mathbf{I}$ as input and outputs local image patch embeddings $\mathbf{F}=\{\mathbf{f}_{n}\}_{n=1}^{N}\in\mathbb{R}^{N\times d'}$, where $N$ is the number of patches. The operations are formally outlined as:
\begin{equation}
    \mathbf{g}_i = \mathcal{E}_{I}(\mathbf{I}_{i}),  \quad
    \mathbf{H}_{i} = \mathcal{E}_{L}(\mathbf{L}_{i}), \quad
    \mathbf{F}_{i} = E(\mathbf{I}_{i}).
    \label{eqn:overview}
\end{equation}
\noindent To aggregate the embeddings of the local patch image $\mathbf{F}$ into a single global feature $\mathbf{f}^{id}\in\mathbb{R}^{d'}$, we employ a multilayer perceptron (MLP) with a single hidden layer. 
In person ReID, the similarity between two images is determined using the cosine similarity of their respective features $\mathbf{f}^{id}$ whereas the inference process in our ReID system solely relies on the student image encoder $E$, without the need for any additional modules.

\Paragraph{Linguistic Body Shape Description Labeling}
We automate the creation of linguistic descriptors using the CLIP model's ability to interpret images and generate relevant body shape labels. Our descriptors include $M=16$ pairs (\emph{e.g.}, Muscular-Slender, Long Torso-Short Torso, and High-Waisted-Low Waisted) of phrases that effectively contrast body shapes, ensuring robustness against variations in distance, clothing, and camera angles.

\Paragraph{Dual Distillation from CLIP}
CLIP3DReID employs a dual distillation approach of the text and image components of the CLIP model. This involves aligning the student encoder's visual features with the CLIP-generated linguistic descriptions using optimal transport. This alignment optimizes the learning process, enabling the student encoder to internalize domain-invariant features that are critical for consistent recognition performance under diverse conditions.

\Paragraph{3D Reconstruction Regularization}
As shown in Fig.~\ref{fig:body_overview}, we incorporate a novel 3D reconstruction regularization using synthetic body shapes derived from the SMPL model. This technique emphasizes learning invariant features across different domains, significantly boosting the generalizability of our model. Synthetic mesh images, along with their generated linguistic descriptors, are used to further refine the model's ability to discern and reconstruct accurate body shapes.

%\subsubsection{Open-set Search \textcolor{red}{Yiyang}}
\subsubsection{Open-Set Search}
%\noindent\textcolor{red}{$<$ \emph {0.5 page, including an illustrative figure}}

\begin{figure}
    \centering
    \includegraphics[width=\linewidth]{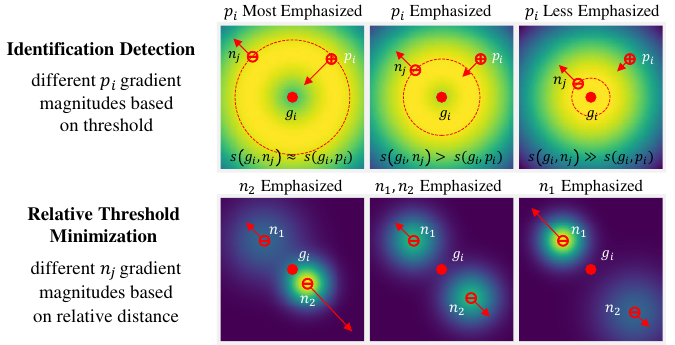}
    \caption{Visualization of the proposed open-set loss~\cite{su2024open}. For $R_{\tau}^{det}$, as shown in the top row, the thresholds are determined by the non-mated sample, \(n_j\). The gradient $\partial \mathcal{L}_{open} / \partial p_i$ has the greatest magnitude when it has a similar distance from the gallery \(g_i\) to \(n_j\).
    For Relative Threshold Minimization, as shown in the bottom row, as non-mated sample \(n_2\) moves away from the gallery, its gradient decreases. While \(n_1\) remains at the same location, its gradient increases because it becomes closer to \(g_i\) than \(n_2\).
    The gradients {\it w.r.t.} genuine scores adapt to non-mated scores, and the gradients {\it w.r.t.} non-mated scores are adapted to other non-mated scores.}
    \label{fig:open_set_overview}
\end{figure}

Open-set biometric recognition poses the challenge of not only correctly identifying known subjects from a gallery but also rejecting probe instances that do not have a mate with any enrolled identity. To address this, we introduce a loss function tailored for open-set recognition~\cite{su2024open} that simulates testing scenarios during training to improve generalization and robustness.
%We design loss functions tailored for open-set person recognition~\cite{su2024open} by simulating testing conditions during training. 
Each training batch is partitioned into gallery and probe subsets. A proportion \(p\%\) of subjects are randomly selected as mated, with exemplars distributed across both gallery and probe sets. The remaining non-mated subjects are assigned to the probe set only. This setup creates realistic open-set training scenarios. We denote the mated probe set as \(\mathcal{P}_\text{K}'\), the non-mated probe set as \(\mathcal{P}_\text{U}'\), and the gallery as \(\mathcal{G}'\).

% Similarity scores are computed using the feature space metric. For Euclidean distance:
% \begin{align}
% \operatorname{s}(p_i, g_j) = \frac{1}{1 + \operatorname{d}(p_i, g_j)},
% \end{align}
% where \(\operatorname{d}(\cdot, \cdot)\) is the Euclidean distance. Cosine similarity is used for cosine-based metrics.

As illustrated in Fig.~\ref{fig:open_set_overview}, we address three types of errors: (1) failing to detect a mated probe with a threshold \(\tau\), (2) failing to identify a mated probe within the top rank-\(r\) positions, and (3) assigning very high similarity scores to non-mated probes.

\Paragraph{(1) Detection} Detection assesses if a pairwise similarity score exceeds a threshold \(\tau\). For a mated probe \(p_i \in \mathcal{P}_\text{K}'\) and its corresponding gallery subject \(g_i \in \mathcal{G}'\):
\begin{align}
    R^{det}_\tau(p_i, g_i) = \sigma_\alpha\left(\operatorname{s}(p_i, g_i) - \tau\right),
\end{align}
where \(\sigma_\alpha(x) = 1 / (1 + \exp(-\alpha x))\) is a Sigmoid function with hyperparameter \(\alpha\). This focuses on the loss for samples near the threshold \(\tau\). The batch-level detection threshold is:
\begin{align}
    R^{\mathit{det}} = \frac{1}{\left\lvert \mathcal{T} \right\rvert}\sum_{\tau \in \mathcal{T}} R^{det}_\tau(p_i, g_i),
\end{align}
where \(\mathcal{T} = \{s(n_j, g_i) \vert n_j \in \mathcal{P}_\text{U}' \}\). The threshold set $\mathcal{T} $ aligns with the FNIR @ FPIR metric.

\Paragraph{(2) Identification} Identification ensures that the mated subject ranks correctly in the gallery. The identification score \(S^{id}(p_i, g_i)\) is:
\begin{align}
    R^{id}(p_i, g_i) = \sigma_\beta(1 - \operatorname{softrank}(p_i, g_i)),
\end{align}
where \(\operatorname{softrank}(p_i, g_i) = \sum_{g_j \in \mathcal{G}'} \sigma_\gamma(\operatorname{s}(p_i, g_j) - \operatorname{s}(p_i, g_i))\). \(\operatorname{softrank}\) reflects \(g_i\)'s rank by summing scores of more similar gallery subjects. The identification-detection loss \(\mathcal{L}_{\operatorname{IDL}}\) is:
\begin{align}
    \mathcal{L}_{\operatorname{IDL}} = -\frac{1}{\left\lvert \mathcal{P}_\text{K}' \right \rvert} \sum_{p_i \in \mathcal{P}_\text{K}'} R^{det}(p_i, g_i) \cdot R^{id}(p_i, g_i).
\end{align}
This loss penalizes failures in detection and identification.

\Paragraph{(3) Relative Threshold Minimization} To reduce false positives, we penalize high non-mated scores using their weighted average:
\begin{align}
    \mathcal{L}_{\text{RTM}} = \frac{1}{\sum_{j=1}^{n} e^{s_j}} \sum_{j=1}^{n} e^{s_j} \cdot s_j,
\end{align}
where \(e^{s_j}\) is the softmax-weighted score. This approach lowers all high scores, promoting generalization.

\Paragraph{Overall Loss} The final loss combines \( \mathcal{L}_{\operatorname{IDL}}\) and \(\mathcal{L}_{\text{RTM}}\) as follows:
\begin{align}
    \mathcal{L}_{open} = \mathcal{L}_{\operatorname{IDL}} + \lambda \cdot \mathcal{L}_{\text{RTM}},
\end{align}
where \(\lambda\) controls the trade-off. This formulation aligns optimization with open-set evaluation, reducing threshold values and leveraging non-mated score magnitudes for robust feature learning.

To optimize the model to distinguish between close-range data in the gallery and long-range data in the probe during evaluation, we modify the triplet loss as follows. In the standard triplet loss, both close-range and long-range data can serve as anchors, positives, and negatives. We adjust this by restricting close-range data to serve only as anchors, while long-range data is used exclusively as positive and negative samples.

%--------------------------------------------------------
\begin{figure}[t]
    \centering
    \includegraphics[width=1\linewidth]{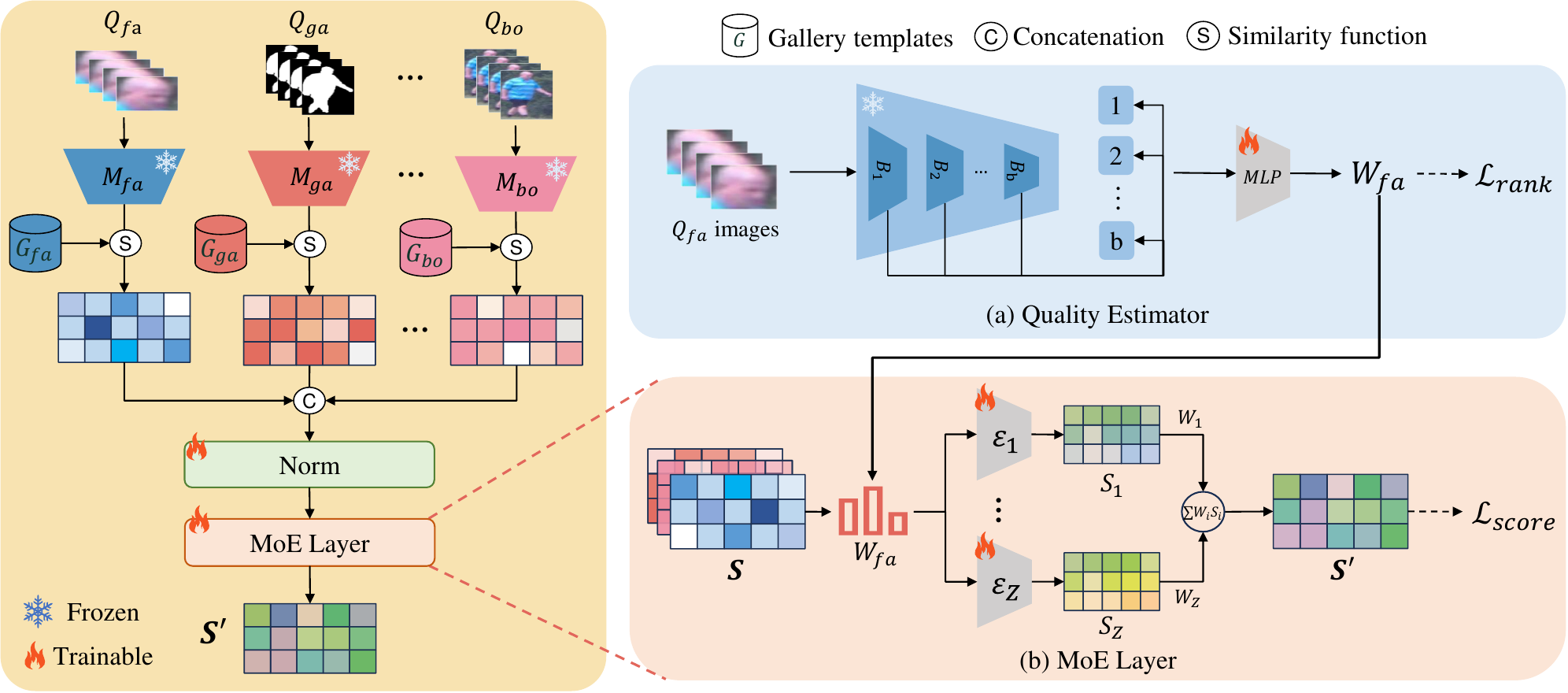}
    \caption{The architecture of the quality-guided mixture of score-fusion experts includes a $\mathit{Norm}$ layer and an $\mathit{MoE}$ layer to process concatenated score matrix $\mathbf{S}$ from the model set $\{\mathcal{M}_{fa}, \mathcal{M}_{ga}, \dots, \mathcal{M}_{bo}\}$. The $\mathit{MoE}$ layer contains experts $\{\varepsilon_1, \varepsilon_2, \dots, \varepsilon_Z\}$ to individually encode the fused score matrices. A quality estimator (QE) uses the intermediate feature $\mathcal{I}_{fa}$ to generate weights $W_{fa}$, which control score-fusion experts for a weighted sum, producing the final fused score matrix $\mathbf{S}'$.}
    \label{fig:LSF_MoE}
\end{figure}
%--------------------------------------------------------

%----------------------------------------------------
\begin{table*}[t]
\centering
\resizebox{0.8\linewidth}{!}{
\begin{tabular}{@{}lcccccc@{}}
\toprule
Dataset      & \# Subjects & \# Media (videos/images) & Max. range  & Max. elevation & Clothing change  \\ \midrule \midrule
%IJB-MDF      & 251         & 500 m       & Ground         & No                & Yes    \\
%IJB-S        & 202         & Estimated $\leq$500 m & FAA limit       & Yes (40)          & Yes    \\
%RPField      & 112 + 4,000 distractors & 158 m & Cameras on poles & No                & Yes    \\
%PRAI-1581    & 1,581       & 60 m        & Ground         & No                & No     \\
%MEVA         & 100         & Estimated $\leq$500 m & FAA limit       & No                & Yes    \\
%UG2+         & N/A         & Estimated $\leq$500 m & FAA limit       & No                & No     \\
%UCCS         & 1,732       & 150 m       & Ground         & Yes               & Yes    \\
%DukeMTMC     & 2,700       & Estimated $<$100 m & Estimated $<$10 m & No                & Yes    \\
%P-DESTRE     & 269         & 6.7 m       & Ground         & Yes               & Yes    \\
%LFRID        & 100 (+ 49 sequestered) & 1000m & Ground  & No                & No     \\
%BGC1   & 312 (+ 161 distractors) & \textcolor{red}{?} & 500 m & 50$^{\circ}$  & Yes                    \\
%BGC2   & 302 (+ 280 distractors) & \textcolor{red}{?} & 1000 m & 50$^{\circ}$ & Yes                    \\
%\hline
BRIAR-BRC & 995 & 134,758/339,190 & 1000 m & 50$^{\circ}$  & Yes                  \\
MSU-BRC  & 452 & 6,039/17,563 & 1000 m & 50$^{\circ}$  & Yes                  \\
Accenture-BRC  & 512 & 20,506/21,204 & 920 m & 45$^{\circ}$  & Yes                 \\ %24215+18937 
Kitware-BRC  & 509 & 24,588/252,187 & 1000 m & 43$^{\circ}$  & Yes                   \\
USC-BRC  & 290 & 16,509/10,194 & 600 m & 50$^{\circ}$ & Yes                   \\
STR-BRC  & 436 &8,394/25,134 & 500 m & 45$^{\circ}$    & Yes                    \\ \hline
Total  & 3,194 & 210,794/665,472  & 1000 m & 50$^{\circ}$ & Yes\\
\bottomrule
\end{tabular}}
\caption{Overview of BRIAR Research Collection (BRC) training datasets, including the government collections training set (BRIAR-BRC) and contributions from five different BRIAR performer teams (MSU, Accenture, Kitware, USC, and STR).}
\label{tab:dataset_comparison}
\end{table*}
%----------------------------------------------------
%----------------------------------------------------
\begin{table}[t]
    \centering
    \resizebox{0.43\textwidth}{!}{%
    \begin{tabular}{|r|r|r|r|r|}
    \hline
        Probe Set & Subjects & Vid. Tracks & Videos & Frames \\ \hline
        All Probes & 424 & 10,731 & 7,260 & 2,396,734 \\ \hline
        Control Probes & 424 & 8,752 & 5,921 & 2,034,524 \\ \hline
        Treatment Probes & 424 & 1,619 & 1,339 & 362,210 \\ \hline
    \end{tabular}
    }
    \vspace{-2mm}
    \caption*{(a) Probe statistics}
    \vspace{2mm}
    \resizebox{0.48\textwidth}{!}{%
    \begin{tabular}{|r|r|r|r|r|r|r|}
    \hline
        Simple & Subjects & Distract & Stills & Vid. Tracks & Videos & Frames \\ \hline
        Gallery1 & 209 & 674 & 79,480 & 10,499 & 10,499 & 5,901,117 \\ \hline
        Gallery2 & 215 & 669 & 79,566 & 10,621 & 10,621 & 5,951,395 \\ \hline
        Unique IDs & 424 & 675 & 99,007 & 12,264 & 12,264 & 6,975,748 \\ \hline
    \end{tabular}
    }
    \vspace{-2mm}
    \caption*{(b) Simple gallery statistics}
    \vspace{2mm}
    \resizebox{0.48\textwidth}{!}{%
    \begin{tabular}{|r|r|r|r|r|r|r|}
    \hline
        Blended & Subjects & Distract & Stills & Vid. Tracks & Videos & Frames \\ \hline
        Gallery 1 & 214 & 367 & 32,673 & 11,961 & 6,535 & 4,430,353 \\ \hline
        Gallery 2 & 210 & 327 & 30,699 & 10,486 & 5,490 & 3,889,257 \\ \hline
        Unique IDs & 424 & 679 & 62,382 & 22,134 & 11,876 & 8,214,485 \\ \hline
    \end{tabular}
    }
        \vspace{-2mm}
    \caption*{(c) Blended gallery statistics}
    \vspace{1mm}
    \caption{BRIAR V5.0.1 evaluation protocol: probe and gallery statistics. While the Blended gallery was originally intended to be more difficult, in V5.0.1 it often yields higher performance because it includes high-quality mugshot-like crops. In contrast, the Simple gallery better reflects unconstrained real-world enrollments.}
    \label{tab:evp501}
\end{table}
%----------------------------------------------------

\subsection{Quality-Guided Multi-Modal Fusion}
As illustrated in Fig.~\ref{fig:LSF_MoE}, our fusion module leverages a learnable Mixture-of-Experts (MoE) mechanism guided by modality-specific quality scores.
Given a probe feature $p_{fa}\in \mathbb{R}^{d_{fa}}$ from probe set \(\mathcal{P}_{fa}\), where $fa$ is the face modality and $d_{fa}$ is the feature dimension, we follow~\cite{kim2022cluster} to extract intermediate features $\mathcal{I}_{fa}$ from the backbone and then feed into an encoder to predict the quality weight $W_{fa} \in \mathbb{R}$ produced by the sigmoid function. We design an MoE layer (see Fig.~\ref{fig:LSF_MoE}) with multiple score-fusion experts, controlled by $\mathit{\mathcal{N}_r}$ that learns to perform score-fusion based on quality weights. Given $W_{fa}$ as the quality weight and $\varepsilon_{fa}$ controlled by $W_{fa}$, we aim for expert $\varepsilon_{fa}$ to prioritize facial modality when $W_{fa}$ is high. Conversely, when $W_{fa}$ is low, another expert, $\varepsilon_j$ (controlled by $1 - W_{fa}$), shifts focus to other modalities, reducing reliance on the face. This approach ensures that higher-quality modalities have a greater influence on the output, while lower-quality ones contribute less, optimizing overall performance.

 As illustrated on the left side of Fig.~\ref{fig:LSF_MoE}, for a query feature image, we generate the input score matrix $\mathbf{S}=\{s_{fa}, s_{ga}, \dots, s_{bo}\} \in \mathbb{R}^{N_G \times N_M}$ from model set, respectively, where $ga=$ is the gait modality and $bo$ is the body shape modality. $N_G$ is the number of gallery features and $N_M$ is the number of models ($N_M=3$ in our case). The final fused score matrix $\mathbf{S}'$ is computed as a weighted sum of the outputs from all experts: $\mathbf{S}'=\sum W_z \mathbf{S}_z$, where $\mathbf{S}_z$ is the output score matrix from $\varepsilon_z$. By using quality weights to modulate $\mathbf{S}'$, each expert learns how the contributions of different modalities’ scores to $\mathbf{S}'$ should be adjusted in response to changes in their quality levels. We employ a two-stage training: (1) training QE with proposed ranking loss and (2) freezing QE while training the learnable score-fusion model with score triplet loss $\mathcal{L}_{\mathit{score~}}$:
\begin{equation}
    \mathcal{L}_{\mathit{score~}}=\operatorname{ReLU}(\mathbf{S}'_\mathit{nm})+\operatorname{ReLU}(m-\mathbf{S}'_\mathit{mat}),
    \label{eq:score_loss}
\end{equation}
where $\mathbf{S}'_\mathit{nm}$ is the non-match scores of $\mathbf{S}'$, $\mathbf{S}'_\mathit{mat}$ is the match score of $\mathbf{S}'$, and $m$ is the margin value. By further constraining the boundary of non-match scores, the model learns to widen the gap between match and non-match scores while simultaneously reducing the value of non-match scores.

%% file: sec/sec_4_exp.tex
%---------------------------------------------------
\begin{figure*}[t]
\centering
\includegraphics[width=1.0\linewidth]{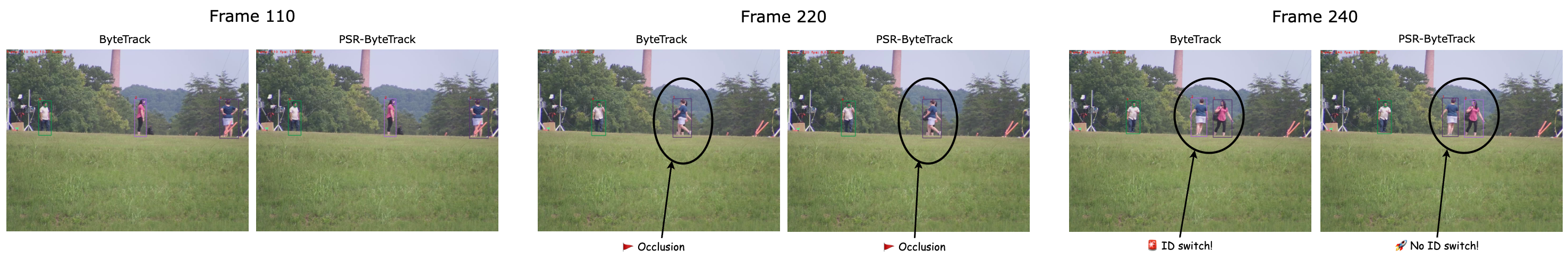}
\caption{Comparison of tracking performance before and after applying PSR-ByteTrack. In an earlier frame (frame 110), we can see that there are three subjects in the probe. After an occlusion (frame 220), it is evident in frame 240 that ByteTrack suffers from the problem of ID-switch. However, our PSR + ByteTrack tracker is able to correctly associate bounding boxes to the appropriate subjects, thereby mitigating the problem of ID-switch.}
\label{fig:gt-psrv2-frames}
\end{figure*}

\input{sec/tables/optim_det}
%------------------------------------------------------

%\section{Experiments  \textcolor{red}{All}}\label{sec:exp}
\section{Experiments}\label{sec:exp}
All experiments are conducted within a configurable containerized environment using PyTorch 2.2.2. We utilize 8 NVIDIA RTX A6000 GPUs (48 GiB VRAM each), deployed across two dual-socket servers equipped with either AMD EPYC 7713 64-Core or Intel Xeon Silver 4314 32-Core processors. %Each experiment runs for 48 hours on average.
%All modules are run together in a configurable container environment on PyTorch version 2.2.2. We perform experiments on $8$ Nvidia RTX A6000s, with $48$ GiB of VRAM, over the course of $48$ hours on $2$ dual-socket servers with either AMD EPYC $7713$ $64$-Core or Intel Xeon Silver $4314$ $32$-Core processors.

%%%---------------
\Paragraph{BRIAR Datasets and Protocols}
We conduct experiments using the complete IARPA BRIAR dataset~\cite{BRIAR}, which includes all five Biometric Government Collections (BGC1–5). These collections span a wide range of conditions—varying distances (up to 1000 meters), elevated viewpoints (up to 50$^\circ$), and diverse environments (urban, semi-structured, and open-field)—making them well-suited for evaluating unconstrained whole-body biometric recognition. 
In addition to the government-collected data, the BRIAR dataset incorporates training data contributions from five BRIAR performer teams at their individual locations: Accenture, Kitware, MSU, USC, and STR. Each BGC collection is partitioned into BRC (training) and BTC (testing) subsets. Tab.~\ref{tab:dataset_comparison} summarizes the training data across all six sources, comprising 3,194 unique subjects in total.

\noindent\emph{\textbf{Training data:}} The feature encoding models for face, gait, and body shape are trained using distinct datasets tailored for each modality:

$\diamond$ \emph{Face Models}: As detailed in Tab.~\ref{tab:dataset_comparison}, training utilizes BRS subsets from all BGC collections, encompassing data from unique $3,194$ subjects across millions of images and video frames. We further augment this set using the WebFace12M dataset~\cite{zhu2021webface260m}.

$\diamond$ \emph{Gait and Body Shape Models}: In addition to the BRS subset from all BGC collections, training for gait and body models integrates the CCGR~\cite{zou2024cross} and CCPG~\cite{Li_2023_CVPR} datasets. These additional public domain datasets enhance our models' ability to accurately encode gait and body shape features under a variety of real-world conditions.

% $\diamond$ Face Models: As detailed in Tab.~\ref{tab:dataset_comparison}, training capitalizes on the BRS subset from all BGC collections, encompassing data from $3,194$ subjects across millions of images and frames. Additionally, the training dataset is augmented with the WebFace12M dataset~\cite{zhu2021webface260m}.

% $\diamond$ Gait and Body Models: In addition to the comprehensive BRS subset from all BGC collections, training for gait and body models integrates the CCGR~\cite{zou2024cross} and CCPG~\cite{Li_2023_CVPR} datasets. These datasets provide specialized data that enhances our models' ability to accurately encode gait and body shape features under a variety of real-world conditions.

%%
\noindent\emph{\textbf{Testing data:}}  
Our evaluation employs the BRIAR Testing Set
(BTS), aligned with Evaluation Protocol V5.0.1 (EVP 5.0.1\footnote{EVP 5.0.1 includes two gallery configurations: \emph{Simple} and \emph{Blended}. Unless otherwise specified, we report results using the \emph{Simple} gallery setup, which is the standard configuration commonly used in BRIAR evaluations.}) and detailed in Tab.~\ref{tab:evp501}. 
This subset is methodically organized into galleries and probe datasets to fulfill specific roles within our testing framework. The galleries, designed to assess recognition capabilities, consist of two distinct setups: Gallery 1 and Gallery 2. The probe datasets are divided into control and treatment scenarios. The control category includes clips from BGC videos where the face or body identity is most readily identifiable, serving as a benchmark to evaluate baseline algorithm performance. Conversely, the treatment category contains clips where identifying facial or body features are more challenging, reflecting the primary evaluation conditions envisaged by the BRIAR protocol. Each of these categories, control and treatment, is further subdivided into ``face-included'' and ``face-restricted'' scenarios. The face-included scenario focuses on assessing face recognition capabilities, while the face-restricted scenario is used to evaluate body and gait recognition or the performance of multi-modal fusion of all the three biometric modalities.

$\diamond$ Face Included Control: Includes visible faces with at least 20 pixels in head height, captured from ground level at a close range of less than 75 meters.

$\diamond$ Face Included Treated: Includes visible faces with the same pixel requirement, captured from long distances or elevated angles, including UAVs.

$\diamond$ Face Restricted Control: Contains data where faces are occluded, of low resolution, or otherwise unusable, captured from ground level at close range.

$\diamond$ Face Restricted Treated: Similar to the above, but captured from long distances or elevated angles, including UAVs.

For select experiments, we also report results under Evaluation Protocol V4.2.0 (EVP 4.2.0)—a subset of V5.0.1—where evaluation is limited to earlier data releases (\emph{e.g.}, BGC1 and BGC2). This allows for legacy benchmarking and direct comparison with previously published baselines.

\Paragraph{Evaluation Metrics}
Following the BRIAR program target metrics~\cite{bba}, we evaluate our system using:
verification (TAR@0.01\% FAR), closed-set identification (Rank-20 accuracy), and open-set identification (FNIR@1\% FPIR), allowing for a thorough examination of FarSights's performance across various settings.

\Paragraph{Baselines}
For person recognition evaluation, we benchmark our system against multiple baselines to place performance in context. First, we compare current FarSight with the original FarSight system~\cite{liu2024farsight}, referred to as \textbf{FarSight 1.0}, to highlight the improvements introduced in our updated framework. Second, we report independent validation results from the 2025 NIST RTE Face in Video Evaluation (FIVE)~\cite{nistfrte2025}, which provides standardized assessments of face recognition systems using the BRIAR dataset. In this evaluation, our system is compared alongside one other top-performing IARPA BRIAR team and two leading commercial biometric systems in this domain.

%In our evaluation, we benchmark FarSight 2.0 against FarSight~\cite{liu2024farsight}. We also compare our results to those achieved by a conventional approach that employs a unified model of an individual without integrating separate biometric modalities like face, body, and gait. %This comparison underscores the enhancements and contributions of our current system, demonstrating significant advancements in creating comprehensive and synergistic biometric recognition solutions.

% \noindent\textcolor{black}{we utilize established benchmarks for each biometric modality to ensure a comprehensive comparison: For facial recognition, we utilize AdaFace coupled with an average feature aggregation strategy, a popular approach known for its excellent performance [26].
% For gait recognition, we adopt GaitBase [14], a solution known for its efficacy. For body shape modality, we employ CAL [17], a SoTA cloth-changing person re-identification method. These benchmarks provide an excellent basis to fairly evaluate our proposed method.}

\subsection{Evaluation and Analysis}

%\subsubsection{Detection and Tracking \textcolor{red}{Georigia Tech}}
\subsubsection{Detection and Tracking}

%\textbf{Tracker Analysis}: 
\textbf{Effectiveness of PSR-ByteTrack}. 
Our enhancements to the ByteTrack framework\cite{zhang2022bytetrack} yield significant improvements in handling multi-subject probes, specifically in reducing identity switch errors. As depicted in Fig.~\ref{fig:gt-psrv2-frames}, initial tracking at frame 110 shows three distinct subjects. By frame 220, a challenging occlusion occurs with overlapping bounding boxes. 
Consequently, in frame 240, ByteTrack suffers from an ID switch error, whereas our PSR-ByteTrack maintains correct subject-bounding box associations throughout the sequence owing to the appearance-based track ID correction postprocessing.

%-------------------------------------------------------------
\begin{table}[t]
    \centering
    \resizebox{1\linewidth}{!}{
    \small
    \begin{tabular}{l|c|c|c}
    \toprule  
    Restoration Type & \makecell{TAR@\\0.1\% FAR ($\uparrow$)} & \makecell{Rank-\\20 ($\uparrow$)} & \makecell{FNIR@\\1\% FPIR ($\downarrow$)} \\
    \hline\hline
      None   & 62.9\% & 87.3\% & 52.4\% \\
      GRTM  & 63.5\% & 86.5\% & 51.6\% \\
      GRTM $+$ vidcls & 63.6\% & \textbf{87.7\%} & 50.1\% \\
      Co-optimized & \textbf{64.1\%} & 87.4\% & \textbf{49.9\%} \\
      \bottomrule
    \end{tabular}
    }
    %\caption{ Face recognition results on EVP 4.2.0 with different restoration configurations. Performance improves progressively from no restoration to co-optimized restoration with recognition-aware training.}
    \caption{Face recognition results on EVP 4.2.0 reduced protocol using the Face-Included Treatment probe set (7,642 tracks from 367 subjects) and Gallery 1 (184 subjects, 4,970 videos, 77,591 stills, and 490 distractors). Columns report performance for: 1:1 verification (TAR@0.1\% FAR), 1:N closed-set identification (Rank-20), and 1:N open-set identification (FNIR@1\% FPIR). "GRTM" refers to our Gated Recurrent Turbulence Mitigation restoration model, "vidcls" adds a video-based classifier to skip unnecessary restoration, and "Co-optimized" denotes joint training with recognition loss.}
    %\caption{FarSight 2.0 Face Results on EVP 4.2.0 with different restoration configurations. The first line is no restoration, the second line is our physics-based video restoration model GRTM, the third row is the GRTM $+$ video restoration classifier, and the final line is our physics-based video restoration model finetuned with restoration-recognition co-training.}
    \label{tab:restoration}
\end{table}
%-------------------------------------------------------------

%--------------------------------------------------------------------
\begin{table*}[t]
    \centering
\resizebox{1\linewidth}{!}{
    \begin{tabular}{c||p{2.0cm}<{\centering}|p{1.8cm}<{\centering}||p{2.0cm}<{\centering}|p{1.8cm}<{\centering}||p{2.0cm}<{\centering}|p{1.8cm}<{\centering}}
     \toprule  
      \multirow{2}{*}{Method} & \multicolumn{2}{c||}{ \makecell{\textbf{Verification (1:1)} \\ TAR@0.1\% FAR $\uparrow$} } & \multicolumn{2}{c||}{\makecell{\textbf{Rank Retrieval (1:N)} \\ Rank-20, Closed Search  $\uparrow$}}  & \multicolumn{2}{c}{\makecell{\textbf{Open Search (1:N)} \\ FNIR@1\% FPIR $\downarrow$} } \\
      \cline{2-7}
      & FaceRestricted & FaceIncluded & FaceRestricted & FaceIncluded & FaceRestricted & FaceIncluded \\ 
      
     \hline\hline     
 FarSight 1.0~\cite{liu2024farsight} (Face)    & 
 $19.8\%$ & $48.5\%$ & $26.6\%$ & $63.6\%$ & $88.8\%$  & $69.7\%$\\  
 \textbf{FarSight}  (Face)   & 
 $30.7\%$ & $66.4\%$ & $42.9\%$ & $80.0\%$ & $82.5\%$  & $57.1\%$\\   
 \hline  
 FarSight 1.0~\cite{liu2024farsight} (Gait)    & 
 $17.7\%$ & $18.9\%$ & $48.6\%$ & $49.5\%$ & $97.6\%$  & $96.7\%$\\  
 \textbf{FarSight} (Gait)    & 
 $61.2\%$ & $66.3\%$ & $90.6\%$ & $93.2\%$ & $78.3\%$  & $75.9\%$\\   
 \hline  
 FarSight 1.0~\cite{liu2024farsight} (Body shape)    & 
 $18.0\%$ & $19.3\%$ & $50.7\%$ & $54.9\%$ & $98.7\%$  & $98.0\%$\\  
 \textbf{FarSight} (Body shape)    & 
 $47.8\%$ & $55.4\%$ & $79.1\%$ & $82.9\%$ & $86.6\%$  & $83.1\%$\\   
 %\hline  
 %\textbf{FarSight} (Face+Gait)     & $??.??$ & $??.??$  & $??.??$ & $??.??$ & $??.??$ & $??.??$ \\
%\textbf{FarSight} (Face+Body shape)     & $??.??$ & $??.??$  & $??.??$ & $??.??$ & $??.??$ & $??.??$ \\
% \textbf{FarSight} (Gait+Body shape)     & $??.??$ & $??.??$  & $??.??$ & $??.??$ & $??.??$ & $??.??$ \\
  \hline  
  FarSight 1.0~\cite{liu2024farsight}          & $30.9\%$ & $48.7\%$  & $62.0\%$ & $77.7\%$ & $91.1\%$ & $79.2\%$\\ 
   \textbf{FarSight}    & $\textbf{65.0\%}$ & $\textbf{83.1\%}$ & $\textbf{91.0\%}$  & $\textbf{95.5\%}$ & $\textbf{69.3\%}$ & $\textbf{44.9\%}$ \\  
\bottomrule
    \end{tabular}
    }
    %\vspace{-2mm}
        %\caption{Person recognition results on BRIAR EVP 5.0.1 across biometric modalities and fusion settings. }
        \caption{Person recognition results on the BRIAR Evaluation Protocol V5.0.1, comparing \textbf{FarSight} (current system) with our previous system FarSight 1.0 across individual biometric modalities (face, gait, body shape) and their fusion. Last row (FarSight) denotes the fusion of all three modalities using our quality-guided fusion strategy. \emph{FaceIncluded} refers to probe segments where faces are visible ($\geq$20 px in head height), while \emph{FaceRestricted} excludes such segments due to occlusion, distance, or resolution. Results are based on the \emph{Treatment Probe Set} (424 subjects, 1,619 video tracks, 1,339 videos, 362,210 frames) and the \emph{Simple Gallery} configuration (424 subjects, 675 distractors, 99,007 stills, 12,264 tracks, 6,975,748 frames). Metrics represent 1:1 verification (TAR@0.1\% FAR), 1:N closed-set retrieval (Rank-20), and 1:N open-set identification (FNIR@1\% FPIR).}
	\label{tab:bts_results}
 \vspace{-2mm}
\end{table*}
%--------------------------------------------------------------------

\noindent
\textbf{Optimized Throughput during Detection}.
We test our improvements to the pipeline on a single-subject probe with a resolution of $896\times1536$ on a NVIDIA A100-80G hyperplane. As shown in Tab.~\ref{tab:detection-dev-perf}, we observe that our system optimizations have a twofold advantage. Firstly, as expected, we observe the throughput to increase with each iteration of updates. In the case of a single GPU with a batch size of 8, we observe $3.13\times$ speedup after merging redundant pre-processing steps, followed by $5.26\times$ speedup for GPU-based pre-processing. Secondly, we observe that moving the pre-processing to GPU has the added effect of alleviating CPU bottlenecks, thereby enabling almost linear scaling of throughput. Here, linear scaling refers to the linear correlation of the increase in problem size to the increase in throughput, thereby demonstrating the absence of any significant bottlenecks. This not only improves the throughput of the detection-tracking submodule but also frees up CPU cores for other submodules in the FarSight system.

\subsubsection{Turbulence Mitigation and Image Restoration}

We evaluate the effectiveness of our restoration strategy by analyzing its impact on face recognition under atmospheric turbulence, as shown in Tab.~\ref{tab:restoration}. 

\noindent\textbf{Baseline.} Without any restoration, our system processes uncorrected video frames. This yields a TAR@0.1\% FAR of 62.9\%, Rank-20 accuracy of 87.3\%, and FNIR@1\% FPIR of 52.4\%, establishing our baseline.

\noindent\textbf{Physics-based Restoration.} Our Gated Recurrent Turbulence Mitigation (GRTM) model improves two out of three metrics—raising TAR to 63.5\% (from 62.9\%) and reducing FNIR to 51.6\%. Although Rank-20 slightly drops to 86.5\% (from 87.3\%), the verification gain suggests better robustness against turbulence-induced distortions.

\noindent\textbf{Restoration with Selective Activation.} When GRTM is augmented with a video classifier (GRTM + vidcls) which triggers restoration only when deemed beneficial, results further improve to 63.6\% TAR and 50.1\% FNIR, with Rank-20 recovering to 87.7\%.

\noindent\textbf{Co-optimized Restoration.} Our full co-optimization strategy—jointly training restoration with a recognition loss—delivers the best overall performance: TAR@0.1\% FAR reaches 64.1\%, FNIR is reduced to 49.9\%, and Rank-20 is maintained at 87.4\%. These gains confirm that task-aware restoration not only enhances image quality but also preserves critical biometric features by avoiding hallucinated details common in purely perceptual models.

\noindent\textbf{Implementation Scope.} To manage computational cost, restoration is applied only to padded face crops, not full video frames. This strategy ensures focus on the most identity-informative regions while maintaining runtime efficiency. Although this analysis targets face recognition, the co-optimization framework is generalizable to other modalities if needed.

\subsubsection{Person Recognition Performance}
The following results are based on the complete FarSight system, incorporating all key modules including open-set search and Quality-Guided Multi-Modal Fusion. Each biometric modality—face, gait, and body shape—shows substantial performance gains over the previous system (\emph{i.e.}, FarSight 1.0~\cite{liu2024farsight}). We evaluate their individual contributions using the BRIAR Evaluation Protocol v5.0.1, and summarize the findings in Tab.~\ref{tab:bts_results}.

\Paragraph{Face}
The updated face feature encoding module achieves significant improvements across all metrics. Compared to FarSight 1.0~\cite{liu2024farsight} on the Face-Included Treatment set of EVP 5.0.1, the proposed FarSight improves verification TAR@0.1\% FAR from 48.5\% to 66.4\%, Rank-20 identification from 63.6\% to 80.0\%, and open-set performance with FNIR@1\% FPIR dropping from 69.7\% to 57.1\%. These gains reflect the impact of the KP-RPE-enhanced vision transformer~\cite{kim2024keypoint} and our recognition-aware restoration module.

\Paragraph{Gait}
The gait feature encoding module exhibits the most substantial improvement on the Face-Included Treatment set, driven by the introduction of the BigGait~\cite{ye2024biggait} model. Verification performance improves from 18.9\% to 66.3\% (TAR@0.1\% FAR), Rank-20 identification rises from 49.5\% to 93.2\%, and FNIR@1\% FPIR decreases from 96.7\% to 75.9\%. These results reflect the model’s enhanced capacity to extract robust gait features using large vision models, particularly under challenging cross-domain conditions.

\Paragraph{Body Shape}
The body shape feature encoding module also demonstrates strong gains on the Face-Included Treatment set. Verification (TAR@0.1\% FAR) improves from 19.3\% to 55.4\%, while Rank-20 identification increases from 54.9\% to 82.9\%. FNIR@1\% FPIR drops significantly from 98.0\% to 83.1\%, indicating improved reliability in open-set scenarios. These improvements are largely attributed to the CLIP3DReID~\cite{liu2024distilling} model, which fuses linguistic cues and visual representations with 3D-aware supervision for enhanced body feature learning.

\Paragraph{Multi-Modal Fusion} While each modality shows marked individual improvements, their complementary strengths become more pronounced when fused. In our full-system setting, FarSight achieves 83.1\% TAR@0.1\% FAR, 95.5\% Rank-20 accuracy, and 44.9\% FNIR@1\% FPIR, outperforming the original FarSight 1.0 by significant margins on the Face-Included Treatment set.

\begin{figure}
\centering
\includegraphics[width=1.0\linewidth]{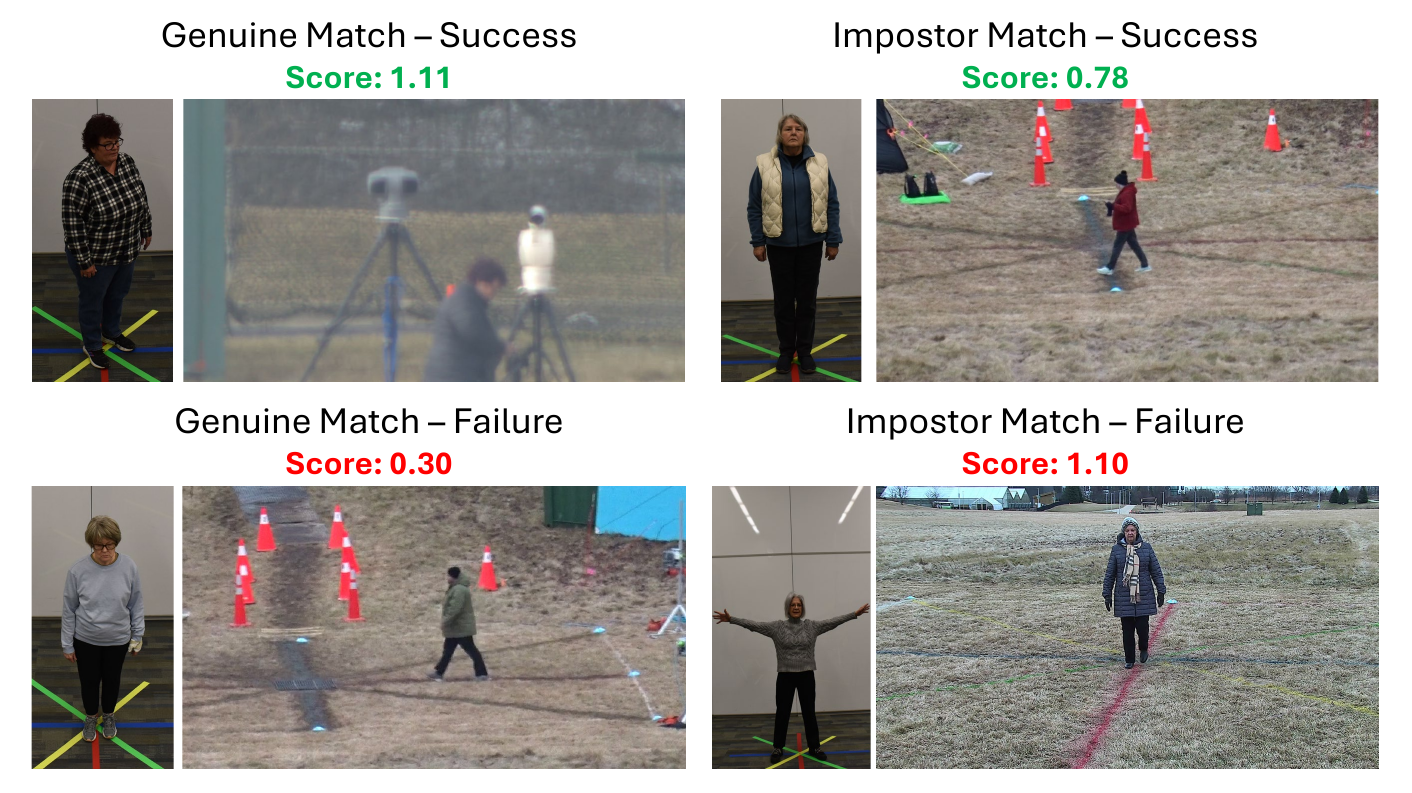}
\caption{Examples of success and failure in closed-set verification. Each pair shows a probe image (right) and its matched gallery image (left), along with the similarity score. Matches are evaluated against a threshold of 0.79, corresponding to 0.1\% False Acceptance Rate (FAR). \emph{Images shown with subject permission for publication.}}
\label{fig:closed_set}
\end{figure}

\begin{figure}
\centering
\includegraphics[width=1.0\linewidth]{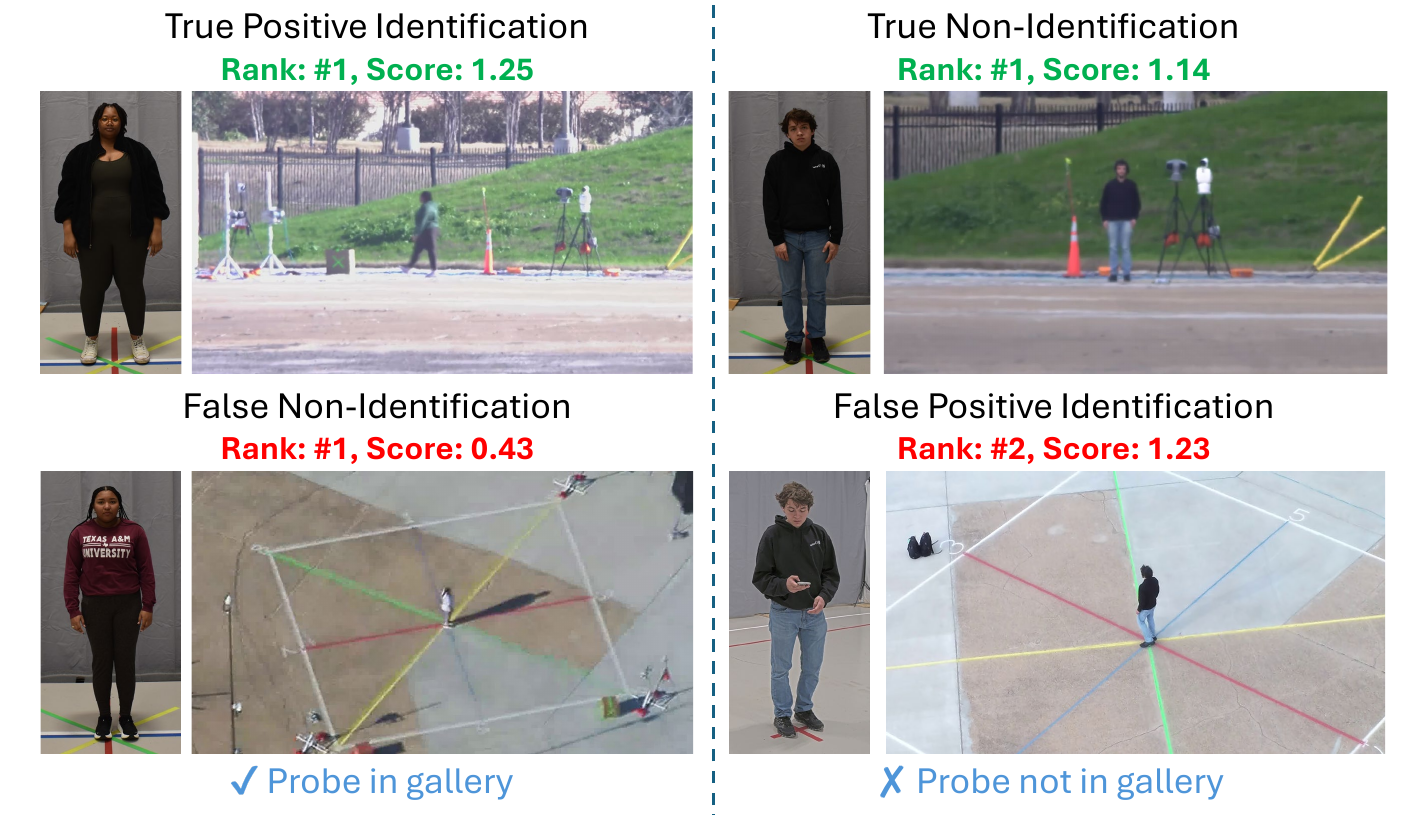}
\caption{Examples of success and failure in open-set identification. Each pair includes a probe image (right) and the top-ranked gallery match (left), along with the match rank and similarity score. An open-set threshold of 1.16 is used to separate accepted matches from rejections. \emph{Images shown with subject permission for publication.}}
\label{fig:open_set}
\end{figure}

\Paragraph{Illustrative Examples of Search Outcomes}
To further illustrate the strengths and limitations of our system, we present qualitative examples of both closed-set and open-set person recognition outcomes. As shown in Fig.\ref{fig:closed_set} and Fig.\ref{fig:open_set}, we include representative success and failure cases across genuine and impostor matches. These examples demonstrate how the system handles identity matching under varied conditions such as distance, altitude and clothing changes. Notably, successful matches exhibit strong visual similarity and alignment, while failure cases often involve challenging views or ambiguous appearances.

%--------------------- close-set --------------------%
%------------------------------------------------%

%--------------------- open-set --------------------%
%------------------------------------------------%

\Paragraph{Independent Validation: NIST FIVE 2025} 
To assess generalization under standardized testing, FarSight’s performance is independently reported in the 2025 NIST RTE Face in Video Evaluation (FIVE)~\cite{nistfrte2025}. The evaluation is conducted using the EVP 5.0.1 Blended gallery, under a 1:N open-set setting with a single frontal still image per subject in the gallery.
As shown in Tab.~\ref{tab:nist_results}, FarSight achieves the best FNIR@1\% FPIR (32\%), outperforming two commercial systems—Sugawara-2 (66\%) and Azumane-2 (53\%)—as well as STR (54\%), which along with MSU are the two remaining performer teams in Phase 3 of the IARPA BRIAR program. These results, reported directly by NIST, further validate FarSight's robustness under operationally challenging scenarios.

%---------------------------- Tab -----------------------
\begin{table}[t]
    \centering
    \small
    \resizebox{1\linewidth}{!}{
    \begin{tabular}{c|c|c|c|c}
    \toprule 
        & \makecell{Sugawara-2\\ (FIVE)} &  \makecell{Azumane-2\\ (FIVE)} &  \makecell{STR\\ (BRIAR)} &  \makecell{FarSight\\ (BRIAR)} \\ \hline \hline
        \makecell{FNIR@\\1\% FPIR}$\downarrow$ &  66\% & 53\% & 54\% & \textbf{32\%} \\
        \bottomrule
    \end{tabular}
    }
    \vspace{-2mm}
    \caption{FNIR@1\% FPIR results from the 2025 NIST RTE Face in Video Evaluation (FIVE)~\cite{nistfrte2025}, evaluated using the BRIAR EVP 5.0.1 protocol with the \emph{Blended gallery} (424 subjects, 679 distractors, 62,382 stills, 22,134 tracks, 8.2M frames) and the \emph{Treatment probe set} (424 subjects, 1,619 video tracks, 1,339 videos, 362,210 frames). Results are reported for a 1:N open-set identification task with one frontal still per subject enrolled.}
    %\caption{FNIR@1\% FPIR results from the 2025 NIST RTE Face in Video Evaluation (FIVE)~\cite{nistfrte2025}, using the EVP 5.0.1 \emph{Blended gallery} in a 1:N open-set setting with a single frontal still per subject.}
    \label{tab:nist_results}
\end{table}

%---------------------------- Tab -----------------------
\subsubsection{Ablation Study}
To better understand the contribution of individual components in the FarSight system, we conduct ablation experiments focused on two core innovations: (1) open-set loss formulation in the gait module and (2) multi-modal fusion framework.

\Paragraph{Impact of Open-Set Losses in Gait Feature Encoding Module}
We apply our open-set loss formulation only to the gait modality in FarSight.
%, training it to better handle open-set conditions by simulating non-mated subjects during training. 
Tab.~\ref{tab:open_set_losses} compares the performance of the gait module with and without the proposed open-set losses using the EVP 4.2.0 protocol. 
The inclusion of open-set losses leads to measurable improvements in 3 out of 4 evaluation metrics. Most notably, verification performance (TAR@0.1\% FAR) improves by 5.7\% in the face-restricted scenario and by 3.2\% in the face-included scenario. While Rank-20 remains unchanged, FNIR@1\% FPIR decreases by 2.3\%, reflecting improved robustness to unknown identities. These results validate the utility of training with simulated open-set conditions.

\begin{table}[]
    \centering\small
    \resizebox{\linewidth}{!}{
    \begin{NiceTabular}{lccc}
        \toprule
        %\Block{1-2}{Open-Set Losses} & & \ding{55} & \ding{51} \\ 
         \Block{1-2}{Open-Set Losses} & & Not Included & Included \\\midrule\midrule
        \Block{2-1}{TAR@0.1\% FAR} & Face Restricted & 44.4\% & \textbf{50.1\%} \\
         & Face Included & 51.6\% & \textbf{54.8\%} \\ \midrule
        Rank-20 & Face Included & 98.7\% & 98.7\% \\ \midrule
        FNIR@1\% FPIR & Face Included & 79.7\% & \textbf{77.4\%} \\ 
        \bottomrule
    \end{NiceTabular}
    }
    \caption{Effect of open-set loss functions in the gait module on EVP 4.2.0.}
    \label{tab:open_set_losses}
\end{table}

\Paragraph{Evaluating Multi-Modal Fusion Strategies}
We further evaluate the contribution of our proposed Quality-Guided Mixture of Experts (QME) fusion strategy by comparing it with the naive score-level fusion used in FarSight.
% FarSight 1.0~\cite{liu2024farsight}. 
Tab.~\ref{tab:fusion_performance} reports the results on the EVP 4.2.0 reduced protocol across both face-included and face-restricted conditions.
Our QME-based fusion approach improves verification performance (TAR) and open-set robustness (FNIR) over the baseline. Notably, FNIR@1\% FPIR improves by 6.2\% in the face-included setting. These gains highlight the effectiveness of using quality-aware fusion and modality-specific weighting over naive score aggregation.

\begin{table}[t]
\resizebox{\linewidth}{!}{
\begin{tabular}{c|c|c|c|c}
\toprule
\multirow{3}{*}{Method}  & \multicolumn{3}{c|}{Face Incl.}         &  \multicolumn{1}{c}{Face Restr.}       \\
\cmidrule{2-5}
& \makecell{TAR@\\0.1\% FAR} & Rank-20 & \makecell{FNIR@\\1\% FPIR} & \makecell{TAR@\\0.1\% FAR}  \\
\hline\hline
Score-level & 77.8\% & \textbf{98.7\%} & 49.1\% & 56.1\%   \\
QME & \textbf{81.0\%} & 98.5\% &\textbf{42.9\%} & \textbf{56.6\%}   \\
\bottomrule
\end{tabular}
}
\caption{Comparison of multi-modal fusion strategies on EVP 4.2.0. "QME" (Quality-Guided Mixture of Experts) refers to our proposed method that adaptively fuses modality scores based on learned quality weights in FarSight.}
\label{tab:fusion_performance}
\end{table}

\subsubsection{Publicly Trained Version of FarSight}
To promote reproducibility and facilitate broader community engagement, we introduce FarSight Public, a version of our system trained and evaluated solely on publicly available data from the MSU-BRC dataset. This dataset, part of the IARPA BRIAR program, is accessible at\footnote{\url{https://cvlab.cse.msu.edu/project-briar.html}}.
The MSU-BRC dataset contains a total of 452 subjects (Tab.~\ref{tab:dataset_comparison}). For this benchmark, we partitioned the data into a disjoint training and testing setup. The training set consists of 228 subjects from MSU-BRC version 2, while the testing set comprises 109 subjects from MSU-BRC version 1. We define an evaluation protocol named MSU 1.0, which includes 2,496 probe segments derived from 626 probe videos. The gallery contains 1,309 distinct videos and 11,815 still images, with 111 distractor identities included to emulate open-set conditions. To simulate clothing variation, different outfits are used for the probe and gallery media.

Although MSU-BRC is not as diverse or challenging as the full BRIAR dataset, it provides a well-structured and accessible benchmark for external validation. We retrain our entire FarSight system on this training split and evaluate its performance using the defined MSU 1.0 protocol. Tab.~\ref{tab:public_results} summarizes the results on the Face-Included Treatment subset. FarSight Public demonstrates strong performance across modalities, with particularly high accuracy for the body shape and fusion modules.

%---------------------------- Tab -----------------------
\begin{table}[t]
\renewcommand{\arraystretch}{1.2}
    \centering
    \small
    \resizebox{0.95\linewidth}{!}{
    \begin{tabular}{c|c|c|c}
    \toprule
       FaceIncluded & \makecell{TAR@\\0.1\% FAR} & \makecell{Rank-\\5} & \makecell{FNIR@\\1\% FPIR} \\ \hline \hline
        FarSight (Face) & 59.8\% & 76.7\% & 58.7\% \\ \hline
        FarSight (Gait) & 55.3\% & 93.4\% & 68.8\% \\ \hline 
        FarSight (Body shape) & 47.6\% & 80.6\% & 73.6\% \\ \hline
        FarSight & 78.0\% & 96.6\% & 39.6\% \\ \bottomrule
    \end{tabular}
    }
    \vspace{-2mm}
    %\caption{FarSight Public results on the MSU-BRC dataset (Face-Included Treatment subset, MSU 1.0 protocol).}
    \caption{FarSight Public results on the MSU-BRC dataset using the MSU 1.0 protocol (Face-Included Treatment subset). The probe set includes 109 subjects across 2,496 tracks from 626 videos. The gallery consists of 1,309 videos and 11,815 stills with mated samples, plus 111 distractor identities for open-set evaluation. Metrics reflect 1:1 verification (TAR@0.1\% FAR), 1:N closed-set identification (Rank-5), and 1:N open-set identification (FNIR@1\% FPIR). Due to the small gallery size, we report Rank-5 instead of Rank-20.}
    \label{tab:public_results}
\end{table}
%---------------------------- Tab -----------------------

\subsection{System Efficiency}
\Paragraph{Template Size}
The template size refers to the amount of data generated and stored per subject for biometric matching. Tab.~\ref{tab:size} summarizes the storage requirements for each modality in the FarSight system.
\emph{\textbf{(i)} Face:}
For face feature encoding, each template contains a 513-dimensional vector. The first 512 dimensions represent the core identity features, while the final dimension stores a face quality score.
%we extract a 513-dimensional feature vector per template. 
Assuming 32-bit floating-point precision, the raw storage requirement is approximately 0.002 MB.
\emph{\textbf{(ii)} Gait:}
Each gait template contains an 8192-dimensional feature vector, resulting in a raw storage size of 0.031 MB.
\emph{\textbf{(iii)} Body shape:}
The body shape representation is encoded as a 2048-dimensional vector, with a raw storage size of 0.008 MB.
\emph{\textbf{(iv)} Combined:}
When all three modalities—face, gait, and body shape—are successfully enrolled, the total raw feature size is approximately 0.041 MB.
While this raw size reflects the uncompressed data representation, practical deployments often involve additional metadata, indexing structures, and compression mechanisms. To estimate real-world storage requirements, we compute the average on-disk size by dividing the total disk space of a deployed gallery by the number of enrolled templates. This yields an effective template size of 0.041 MB, confirming the system's suitability for scalable deployments.

\begin{table}[t]
    \centering
    \resizebox{0.48\textwidth}{!}{%
    \begin{tabular}{c|c|c|c|c|c}\toprule
                      & Face & Gait & Body shape  & Combined & Effective \\\hline\hline
       \makecell{Template \\Size (MB)} & $0.002$  & $0.031$ & $0.008$ & $0.041$ & $0.041$  \\\bottomrule
    \end{tabular}
    }
    \caption{Template size per modality and combined feature representation. }
    \label{tab:size}
\end{table}

% \begin{table}[t]
%     \centering
%     \resizebox{1\linewidth}{!}{
%     \begin{tabular}{c|c|c|c}\toprule
%         Module & 1080p & 4K & Average Combined\\ \hline \hline
%         Detection \& Tracking & 23.4 & 20.1 & 21.8 \\ \hline
%         Restoration           & 64.2  & 27.0  & 45.6\\ \hline
%         Face                  & 31.6  & 34.9  & 33.3\\ \hline
%         Gait                  & 25.0  & 21.5  & 23.3\\ \hline
%         Body Shape                  & 24.0  & 20.3  & 22.2 \\  \hline
%         FarSight System       & 7.0 & 2.9 & 5.0 \\ \bottomrule
%     \end{tabular}
%     }
%     \vspace{-2mm}
%     \caption{ Module-wise processing speed of FarSight in frames per second (FPS) for 1080p and 4K pixel resolution probe videos. The last row reflects effective throughput when all modules operate in parallel. All benchmarks are conducted on 8 NVIDIA RTX A6000 GPUs (48 GiB VRAM each) using PyTorch 2.2.2 in a containerized environment.}
%     \label{tab:processing_time}
%     \vspace{-4mm}
% \end{table}

\begin{table}[t]
\renewcommand{\arraystretch}{1.2}
    \centering
    \small
    \resizebox{1\linewidth}{!}{
    \begin{tabular}{c|c|c|c|c|c|c}
    \toprule
        Resolution & \makecell{Detection\\ \& Tracking}  \ & Restoration & Face & Gait & \makecell{Body\\ Shape}    & FarSight \\
    \hline\hline
        1080p & 23.4  & 64.2  & 31.6  & 25.0  & 24.0  & 7.0  \\
    \hline
        4K    & 20.1  & 27.0  & 34.9  & 21.5  & 20.3  & 2.9 \\
    \bottomrule
    \end{tabular}
    }
    \vspace{-2mm}
    \caption{Module-wise processing speed of FarSight in frames per second (FPS) for 1080p and 4K resolution probe videos. The last column reflects the effective throughput when all modules operate in parallel. All benchmarks are conducted on 8 NVIDIA RTX A6000 GPUs (48 GiB VRAM each) using PyTorch 2.2.2 in a containerized environment.}
    \label{tab:processing_time}
    \vspace{-4mm}
\end{table}

\Paragraph{Processing Speed}
%
%The speed of our FarSight system, as outlined in Tab.~\ref{tab:processing_time}, is examined under stringent conditions to gauge both the efficiency of individual components and the overall pipeline. This system operates asynchronously and concurrently, similar to the actual deployment conditions. To precisely measure efficiency, the components are assessed in a serialized manner, even though they typically run in parallel.
The speed of our FarSight system, summarized in Tab.~\ref{tab:processing_time}, is evaluated under controlled conditions to measure both module-level and system-wide efficiency. While the system is designed to operate asynchronously and concurrently during deployment, for benchmarking purposes, each component is assessed independently in a serialized manner to isolate performance characteristics.
We conduct this assessment using representative sample videos, encompassing $2400$ frames of $1080$p and $1200$ frames of $4$K video, each set originating from four distinct subjects. 
Restoration is selectively applied to detected facial regions. As a result, frames without detected faces naturally reduce the load on both the restoration and face recognition modules. Furthermore, the restoration module incorporates a lightweight classifier that bypasses unnecessary processing when restoration is deemed unlikely to improve recognition.
%The restoration process is primarily directed towards detected faces, which implies that any instances of undetected faces would naturally lead to reduced restoration and face module processing times. The latest version of restoration models also attempt to predict whether the input would benefit from restoration, and skip the restoration phase when it would not.

%\subsection{Future Research: Video Surveillance in the Next Era}
\subsection{Future Research}

\Paragraph{Video Restoration and Co-Optimization} Building on the success of our co-optimization strategy, we plan to extend it to other modalities (\emph{e.g.}, gait, body shape), explore adaptive balancing of restoration and recognition objectives, and integrate uncertainty estimation to prevent identity hallucination. We also aim to design lightweight, real-time architectures suitable for edge deployment in operational environments.

%\Paragraph{Video Restoration \textcolor{blue}{and Co-Optimization}} During the training of the video restoration classifier, due to the characteristics of the BRIAR dataset, the labels of the data are severely imbalanced, which subsequently reduces the learning capability of the network. Overcoming the training data imbalance will help improve the classifier's generalization performance and boost its positive impact on the system.
%\textcolor{blue}{The success of our restoration-recognition co-optimization strategy paves the way for further exploration in biometric system design. Future research directions include extending our joint optimization paradigm to other biometric modalities (e.g., gait and body shape), as well as exploring adaptive optimization techniques capable of dynamically balancing restoration and recognition objectives based on operational contexts. Moreover, integrating uncertainty quantification within the joint optimization loop could further enhance robustness, allowing restoration models to dynamically assess when identity-critical details are recoverable, versus when conservative restoration is preferable to prevent erroneous feature hallucination.}
%\textcolor{blue}{Additionally, real-world aerial biometric surveillance scenarios increasingly require on-device real-time processing, motivating future efforts towards efficient, lightweight co-optimization networks compatible with edge hardware.}

\Paragraph{Detection and Tracking} We plan to integrate a Tracking Any Point (TAP) model~\cite{doersch2022tap} into the FarSight pipeline. By providing dense motion correspondence across frames, TAP can enhance the modeling of fine-grained spatiotemporal features, particularly benefiting gait analysis under occlusion or rapid motion.
%We plan to refine whole-body biometric recognition by integrating a Tracking Any Point (TAP) model, which provides dense correspondences and rich motion details across video frames. This approach enhances the extraction of spatial and temporal features crucial for accurate gait analysis.   

\Paragraph{Biometric Feature Encoding} To improve video-based person recognition, we plan to propose a new framework that adaptively fuses facial, body shape, appearance, and gait cues. Leveraging a dual-input gating mechanism and a mixture-of-experts design, the system will dynamically prioritize feature streams based on video content, enhancing recognition robustness across diverse scenarios.
%\Paragraph{Biometric feature encoding} 
%We aim to enhance video-based person identification by introducing a novel framework that effectively captures and integrates key biometric features—face, static body shape, appearance, and gait. Utilizing a mixture of experts and a dual-input decision gating network, we aim to adaptively prioritize feature processing based on the video context, significantly improving identification accuracy and adaptability.

%We plan to enhance person identification by integrating face, body shape, and gait information into a single unified model using a large vision foundation model. By employing a large vision foundation model to combine these three types of biometric data, we aim to achieve robust and accurate identification, especially in scenarios where single modality alone may be insufficient.

\Paragraph{Multi-Modal Fusion} We aim to further explore score-level fusion strategies within and across modalities. Specifically, we plan to investigate deep learning-based fusion for individual modalities (\emph{e.g.}, multiple face models), and develop a more general, learnable router network to replace fixed quality-based fusion weights. This approach could improve both face recognition and overall system adaptability.

%\Paragraph{Multi-modal fusion.} Our feature plan includes LSF\_MoE model for a single modality since we have multiple FR models in the pipeline. We would like to further explore the effects of DL-based score fusion for a single modality.  The difference between a single modality and multi-modality is the usage of face quality score. Exploring a more general method to replace the face quality weights (\emph{i.e.}, router network) may have a positive effect  towards face recognition performance.

%% file: sec/tables/optim_det.tex
\begin{table*}[!t]
\centering
\resizebox{0.7\textwidth}{!}{
\begin{tabular}{cccccl}
\toprule
\multicolumn{1}{l}{\multirow{2}{*}{}} & \multicolumn{2}{c}{Single GPU} & \multicolumn{2}{c}{\begin{tabular}[c]{@{}c@{}}8 GPUs\\ (Effective throughput)\end{tabular}} \\
\cmidrule{2-5}
 & \textit{bs = 1} & \textit{bs = 8} & \textit{bs = 1} & \textit{bs = 8} \\
\hline\hline

BPJDet + YOLOv8 & 7.7 & 9.7 & 24.5 & 23.6* \\
\midrule

+ merged preprocessing
% & \texttt{fp32} & 20.3 \textcolor{blue}{(2.63x)} & 26.6 \textcolor{blue}{(2.74x)} & 157.0 \textcolor{blue}{(6.4x)} & 203.0 \textcolor{blue}{(8.28x)} \\
& 21.3 \textcolor{blue}{(2.77x)} & 30.4 \textcolor{blue}{(3.13x)} & 160.0 \textcolor{blue}{(6.53x)} & 230.3 \textcolor{blue}{(9.4x)} \\
\midrule

+ GPU-based preprocessing
% & \texttt{fp32} & 29.4 \textcolor{blue}{(3.81x)} & 40.9 \textcolor{blue}{(4.22x)} & 222.8 \textcolor{blue}{(9.1x)} & 320.3 \textcolor{blue}{(13.6x)} \\
 & 31.1 \textcolor{blue}{(4.03x)} & 51.1 \textcolor{blue}{(5.26x)} & 232.7 \textcolor{blue}{(9.5x)} & 389.1 \textcolor{blue}{(16.5x)} \\
\bottomrule
\end{tabular}
}
%\caption{\gtc{Through careful modifications to the system, we observe a significant increase in throughput. We also observe that our modified GPU-friendly pre-processing pipeline alleviates the CPU bottlenecks, benefitting the other submodules.}}
%\caption{Through progressive system optimizations—first merging redundant preprocessing and then offloading it to the GPU—we observe substantial throughput improvements. The GPU-based preprocessing pipeline not only accelerates inference speed but also alleviates CPU bottlenecks, enabling near-linear scalability across 8 GPUs. \textsuperscript{*}Throughput slightly drops due to CPU contention in baseline implementation.}
\caption{Numbers indicate average frames per second (FPS) achieved when processing single-subject probe videos at resolution 896×1536. Measurements were conducted on NVIDIA A100-80GB GPUs with batch sizes (\textit{bs}) of 1 and 8. Optimizations include merging redundant preprocessing and moving preprocessing to the GPU. The last row shows that GPU-based preprocessing yields up to a 5.26× speedup on a single GPU and a 16.5× speedup across 8 GPUs. \textsuperscript{*}Throughput  for the baseline on 8 GPUs drops slightly due to CPU contention.}
\label{tab:detection-dev-perf}
\end{table*}

%% file: sec/sec_5_conclusion.tex
\section{Conclusion}\label{sec:conclu}
We present \textbf{FarSight}, an end-to-end system for whole-body biometric recognition under long-range, unconstrained conditions. By combining physics-based modeling with deep learning across four integrated modules—including detection, recognition-aware restoration, modality-specific encoding, and quality-guided fusion—FarSight addresses key challenges such as turbulence, pose variation, and open-set recognition.
Evaluated on the BRIAR dataset and independently validated by the 2025 NIST RTE FIVE benchmark, FarSight achieves state-of-the-art performance across verification, closed-set, and open-set tasks. 
Specifically, compared to the preliminary system, our system improves 1:1 verification accuracy (TAR@0.1\% FAR) by 34.1\%, closed-set identification (Rank-20) by 17.8\%, and reduces open-set identification errors (FNIR@1\% FPIR) by 34.3\%.
The system is efficient, meets template size constraints, and includes a reproducible public version trained on released data. FarSight offers a strong foundation for next-generation biometric recognition in real-world applications.

%We develop and prototype an end-to-end whole-body person recognition system, \textbf{FarSight 2.0}. Our solution attempts to overcome hurdles such as low-quality video frames, large yaw and pitch angles, and the domain gap between training and test sets by utilizing the physics of imaging in harmony with deep learning models. This innovative approach has led to superior recognition performance, as demonstrated in tests using the BRIAR dataset. With the far-reaching potential to enhance homeland security and forensic identification, the FarSight system paves the way for the next generation of biometric recognition in challenging scenarios.

\Paragraph{Acknowledgments} This research is based upon work supported in part by the Office of the Director of National Intelligence (ODNI), Intelligence Advanced Research Projects Activity (IARPA), via 2022-21102100004. The views and conclusions contained herein are those of
the authors and should not be interpreted as necessarily representing the official policies, either expressed or implied, of ODNI, IARPA, or the U.S. Government. The U.S. Government is authorized to reproduce and distribute reprints for governmental purposes notwithstanding any copyright annotation therein.

%% file: sec/sec_6_bio.tex
\vspace{-15mm}
\begin{IEEEbiography}
[{\includegraphics[width=1in,height=1.25in,clip,keepaspectratio]{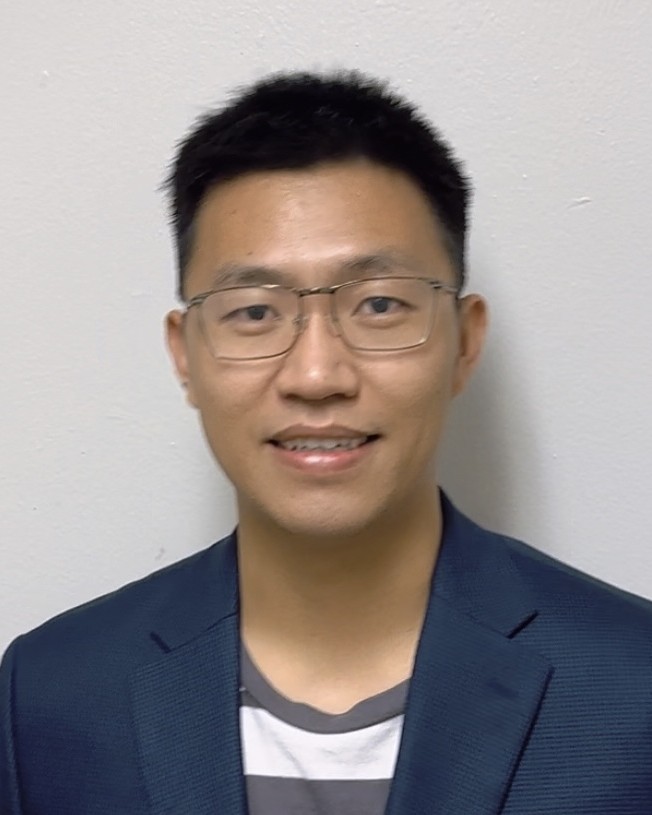}}]{Feng Liu}
(\IEEEmembership{Senior Member,~IEEE}) is an Assistant Professor in the Department of Computer Science at Drexel University. Prior to joining Drexel, he was a postdoctoral researcher at Michigan State University. He received his Ph.D. in Computer Science from Sichuan University. His research spans a wide range of topics in computer vision, machine learning and biometric recognition. 
\end{IEEEbiography}

\vspace{-15mm}
\begin{IEEEbiography}
[{\includegraphics[width=1in,height=1.25in,clip,keepaspectratio]{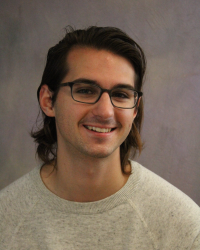}}]{Nicholas Chimitt} (Member, IEEE) is a research scientist at Purdue University, West Lafayette. His research interests include imaging through atmospheric turbulence, wavefront estimation, phase retrieval, and computational optics.
\end{IEEEbiography}

\vspace{-15mm}
\begin{IEEEbiography}
[{\includegraphics[width=1in,height=1.25in,clip,keepaspectratio]{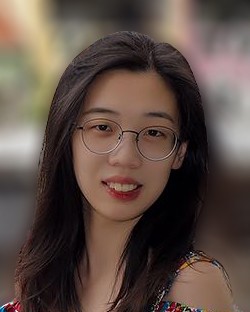}}]{Lanqing Guo} is a postdoc research fellow at The University of Texas at Austin. She earned her Ph.D. in Electrical and Electronic Engineering from Nanyang Technological University, Singapore, where she graduated with the Best Thesis Award. Her research interests include 2D/3D image processing and generation, computational imaging, and computer vision.
\end{IEEEbiography}

\vspace{-15mm}
\begin{IEEEbiography}
[{\includegraphics[width=1in,height=1.25in,clip,keepaspectratio]{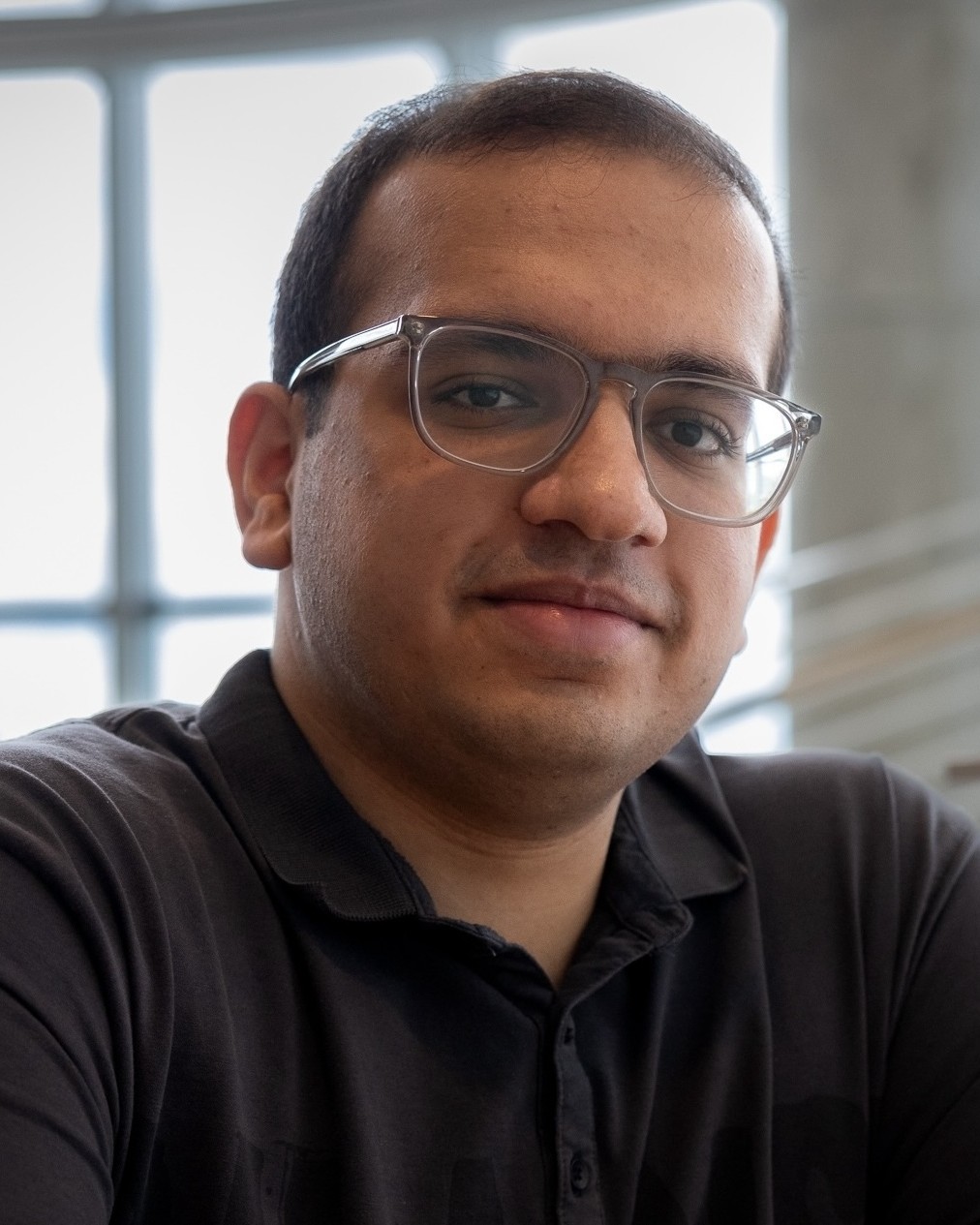}}]{Jitesh Jain} is a Ph.D. student in the School of Interactive Computing at Georgia Tech. He received his Bachelor's in Computer Science and Engineering in 2023 at IIT Roorkee.
His research interests include visual perception and multimodal reasoning.
\end{IEEEbiography}

\vspace{-15mm}
\begin{IEEEbiography}
[{\includegraphics[width=1in,height=1.25in,clip,keepaspectratio]{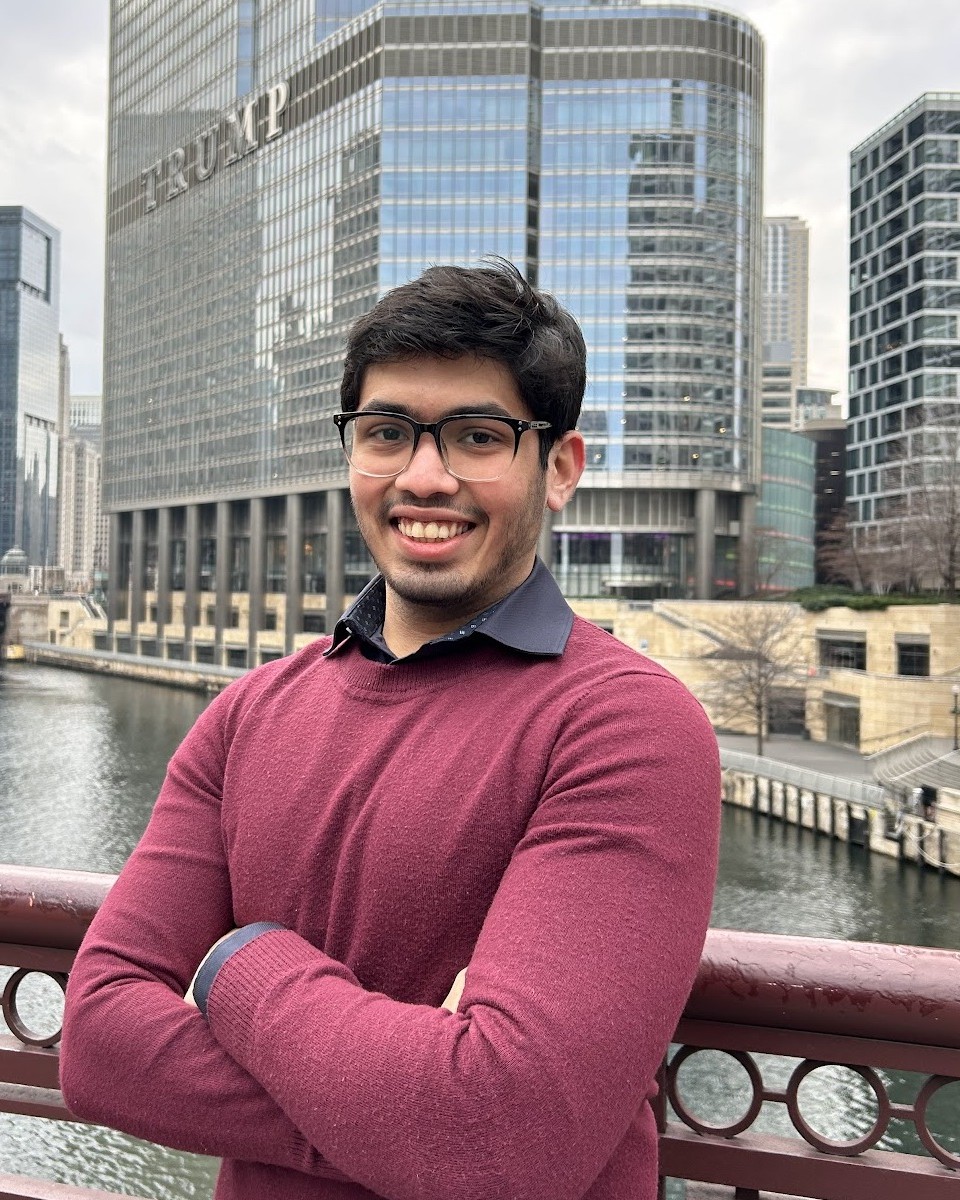}}]{Aditya Kane} is a MS student at Georgia Tech. He completed his Bachelor's in Computer Engineering from Pune Institute of Computer Technology, India.
His research interests include efficiency in deep learning systems.
\end{IEEEbiography}

\vspace{-15mm}
\begin{IEEEbiography}
[{\includegraphics[width=1in,height=1.25in,clip,keepaspectratio]{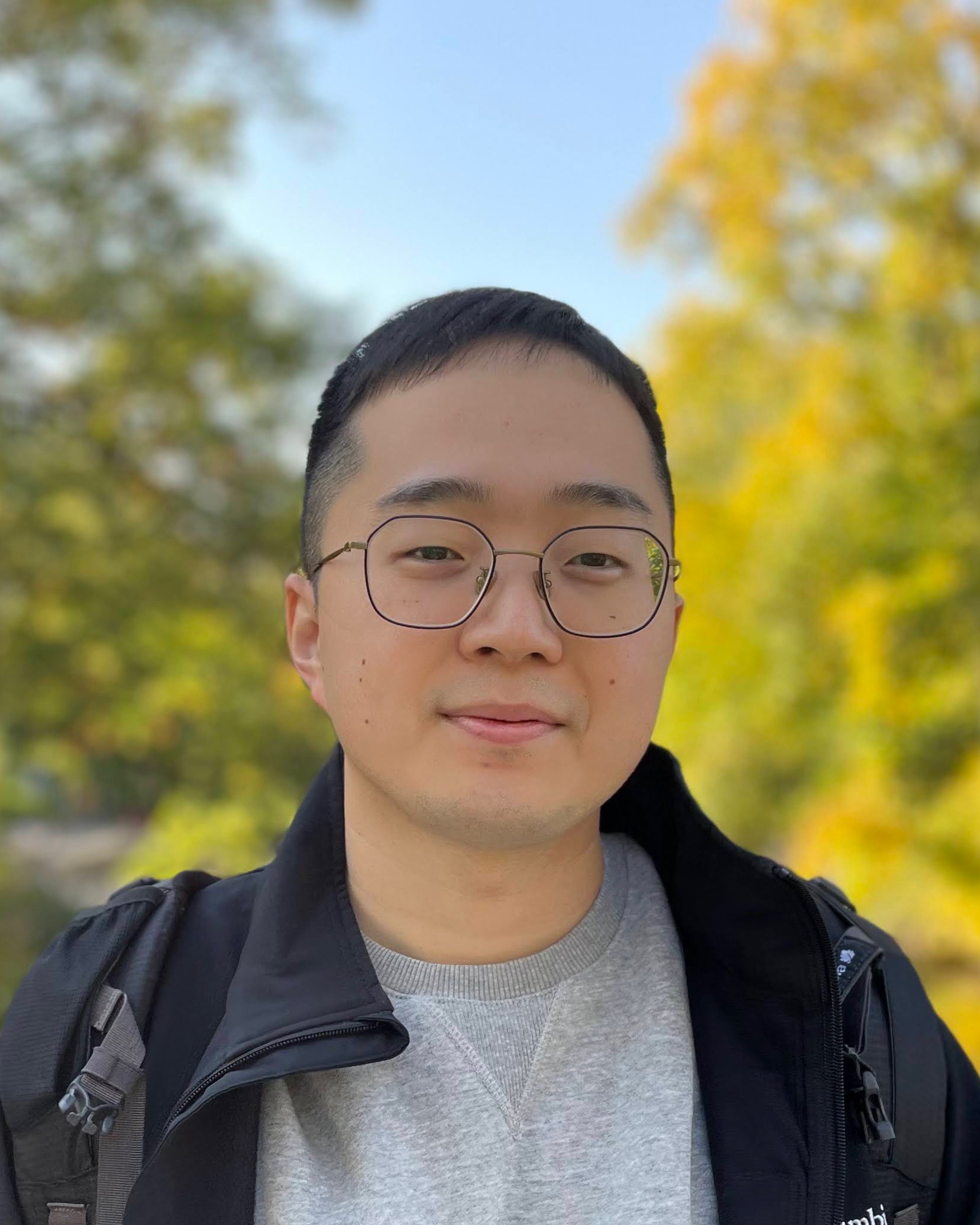}}]{Minchul Kim}
is currently a Ph.D. candidate in the Department of Computer Science and Engineering at Michigan State University. His research interests include face recognition and biometrics.
\end{IEEEbiography}

\vspace{-15mm}
\begin{IEEEbiography}
[{\includegraphics[width=1in,height=1.25in,clip,keepaspectratio]{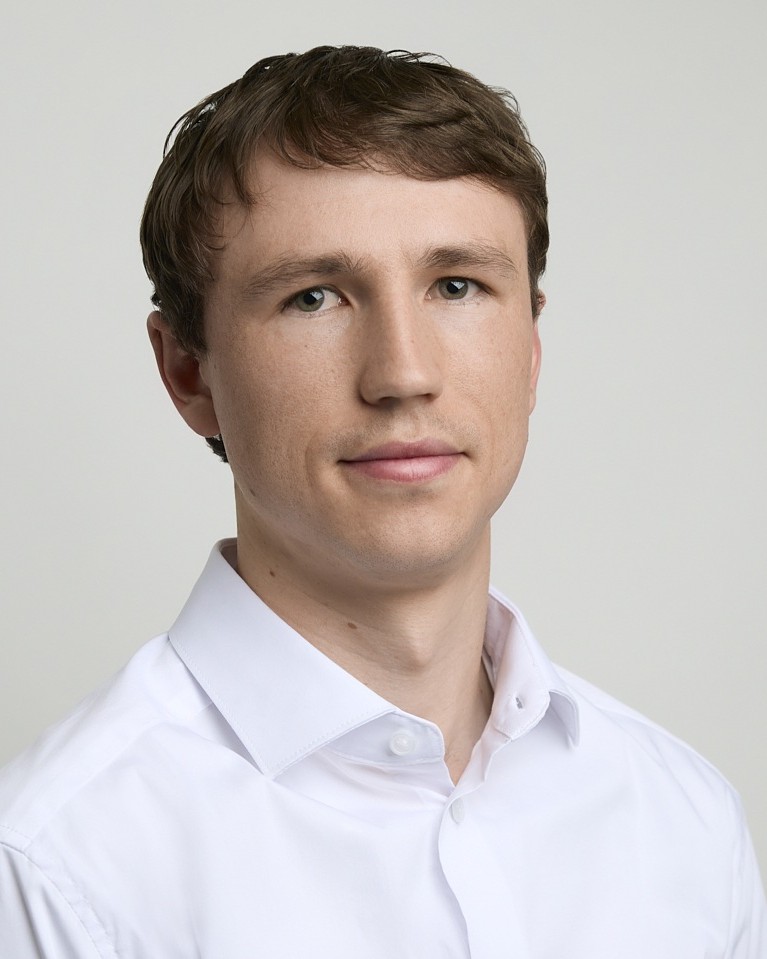}}]{Wes Robbins} is a Ph.D. student in Electrical and Computer Engineering at The University of Texas at Austin. His research interests include computer vision and biometrics.
\end{IEEEbiography}

\vspace{-15mm}
\begin{IEEEbiography}
[{\includegraphics[width=1in,height=1.25in,clip,keepaspectratio]{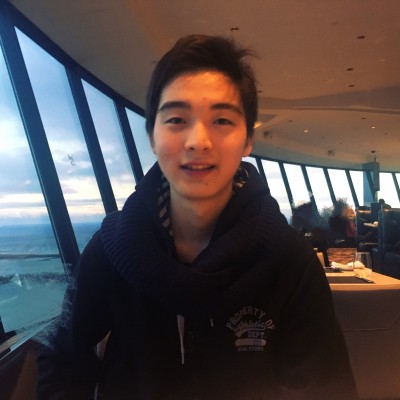}}]{Yiyang Su}
is a Ph.D. student in the Department of Computer Science and Engineering at Michigan State University. He received his B.S. degrees from the University of Rochester. His research interests include biometrics and image generation.
\end{IEEEbiography}

\vspace{-15mm}
\begin{IEEEbiography}
[{\includegraphics[width=1in,height=1.25in,clip,keepaspectratio]{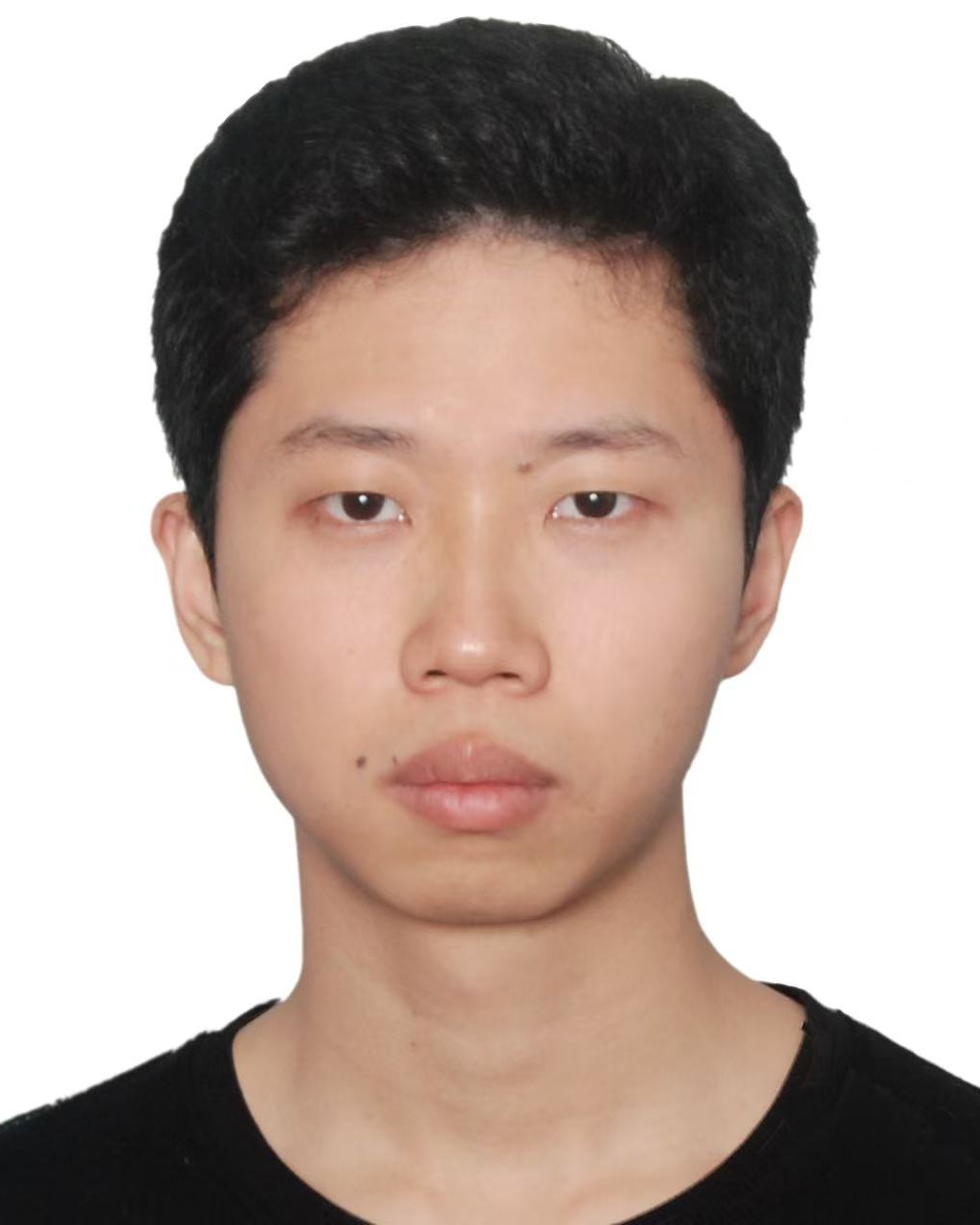}}]{Dingqiang Ye} 
is a visiting scholar in the Department of Computer Science and Engineering at Michigan State University. He is also a master's student in the Department of Computer Science and Engineering at Southern University of Science and Technology. His research interests include computer vision, with a focus on pedestrian analysis, gait recognition, and representation learning.
\end{IEEEbiography}

\vspace{-15mm}
\begin{IEEEbiography}
[{\includegraphics[width=1in,height=1.25in,clip,keepaspectratio]{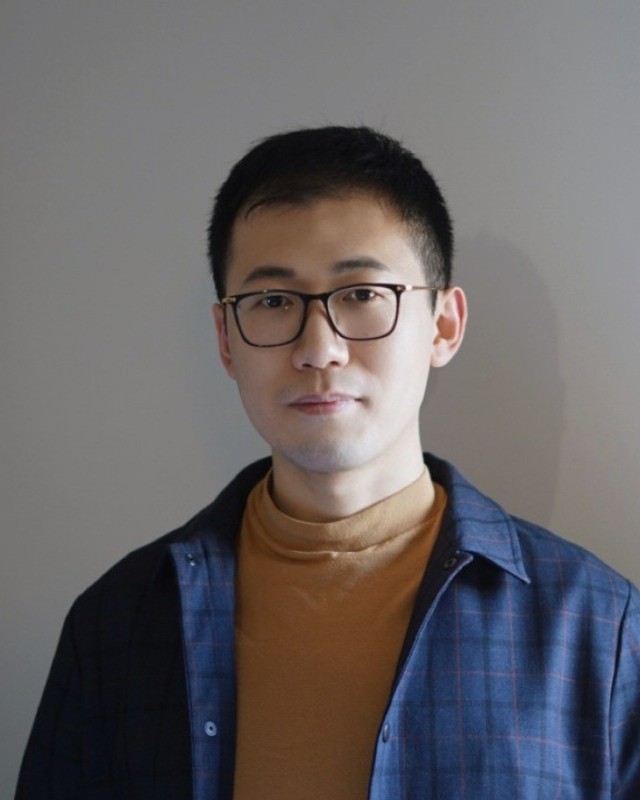}}]{Xingguang Zhang}
%is currently a Ph.D. candidate in the Department of Electrical and Computer Engineering at Purdue University. Prior to that, he received the M.S. degree in ECE from Purdue University and the B.E. degree in Optoelectronic Information Science and Engineering from Zhejiang University, China. His research interests include computational imaging, computer vision, and generative models. Particularly, he is working on atmospheric turbulence mitigation.
(Student Member, IEEE) is a PhD student at Purdue University, West Lafayette. His research interests include computational imaging,  image restoration, and computer vision. 
\end{IEEEbiography}

\vspace{-15mm}
\begin{IEEEbiography}
[{\includegraphics[width=1in,height=1.25in,clip,keepaspectratio]{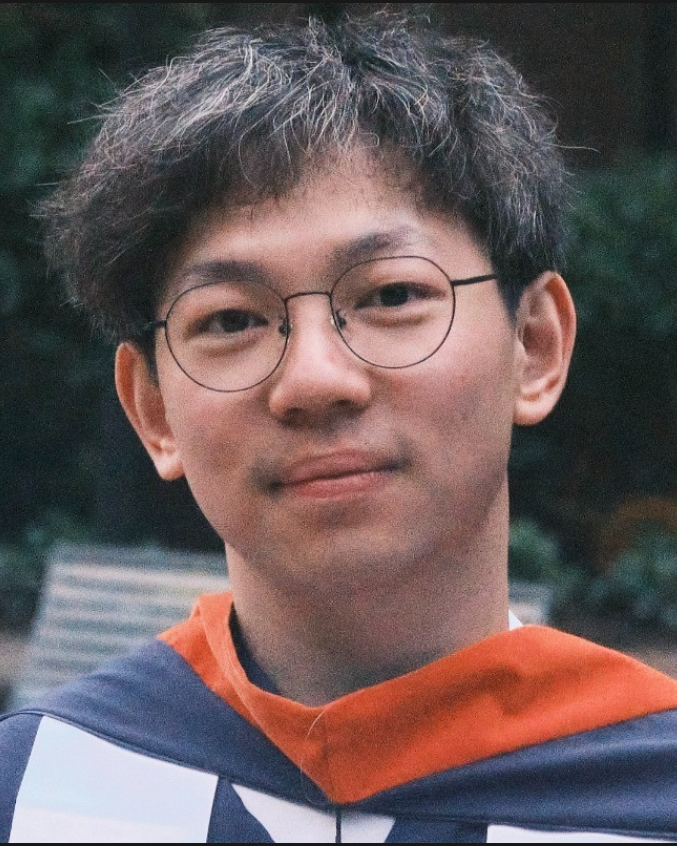}}]{Jie Zhu}
is currently a Ph.D. student in the Department of Computer Science and Engineering at Michigan State University. His research interests include computer vision, with a focus on multi-modal, fine-grained understanding, and representation learning.
\end{IEEEbiography}

\vspace{-15mm}
\begin{IEEEbiography}
[{\includegraphics[width=1in,height=1.25in,clip,keepaspectratio]{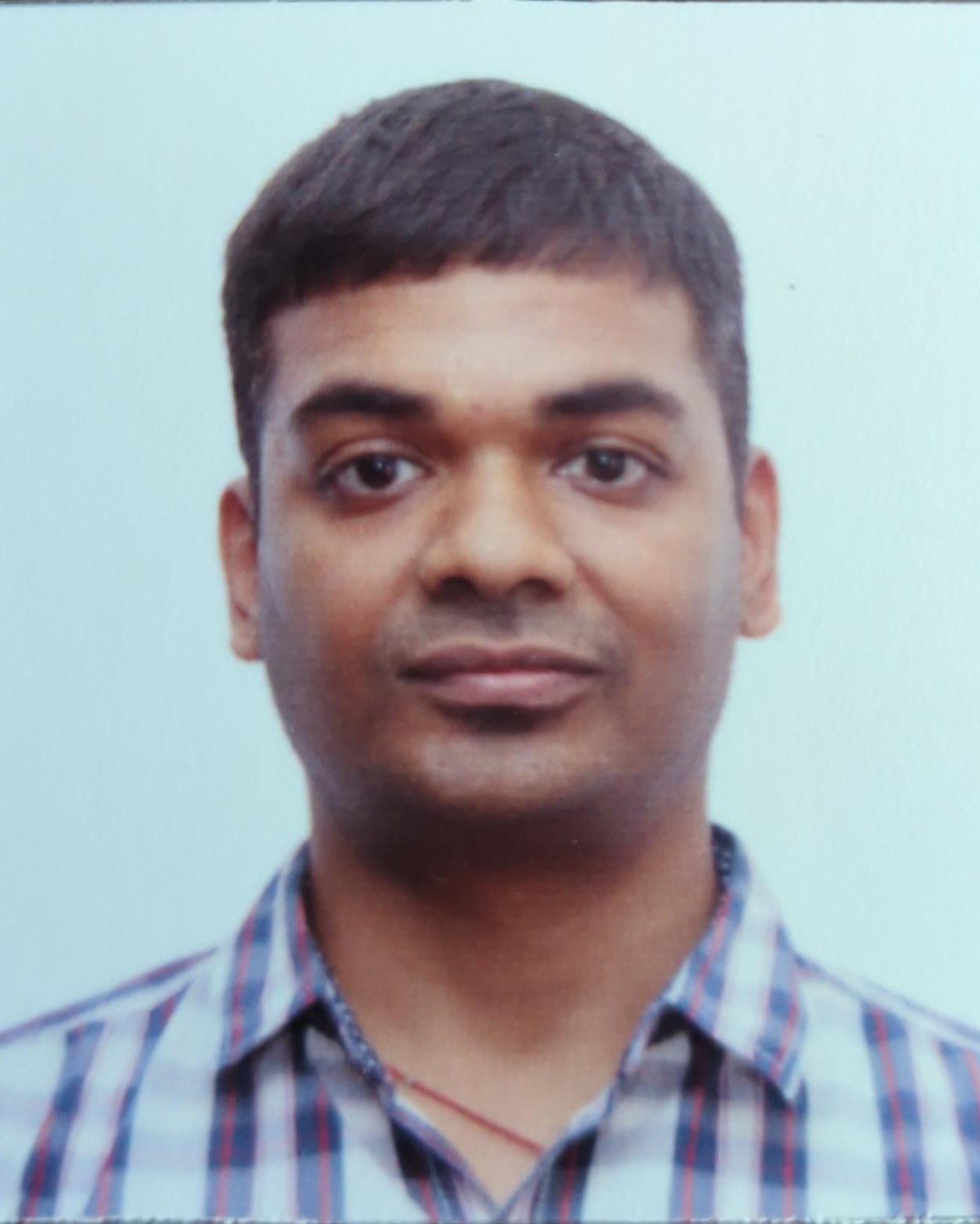}}]{Siddharth Satyakam} is serving as System Analyst II in Department of Computer Science and Engineering in Michigan State University. He received his Bachelor in Electronics and Instrumentation, and his Master in Data Science from SUNY Buffalo. His research interests include biometrics, computer vision, infrastructure as service.
\end{IEEEbiography}

\newpage
%\vspace{-15mm}
\begin{IEEEbiography}
[{\includegraphics[width=1in,height=1.25in,clip,keepaspectratio]{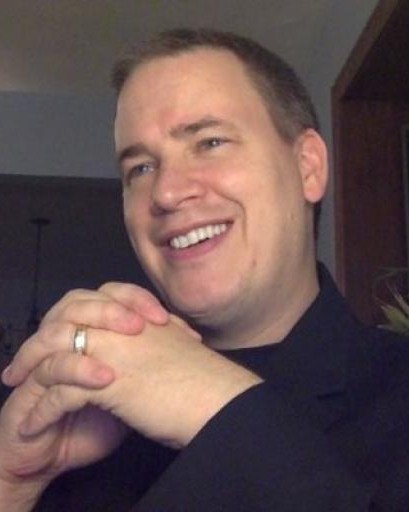}}]{Christopher Perry}
is the BRIAR Lead System Integrator for the Department of Computer Science and Engineering at Michigan State University. He received his Bachelor's in Computer Science and Engineering in 2015 from Michigan State University. His research interests include biometrics, computer vision and distributed/parallel computing.
\end{IEEEbiography}

\vspace{-15mm}
\begin{IEEEbiography}
[{\includegraphics[width=1in,height=1.25in,clip,keepaspectratio]{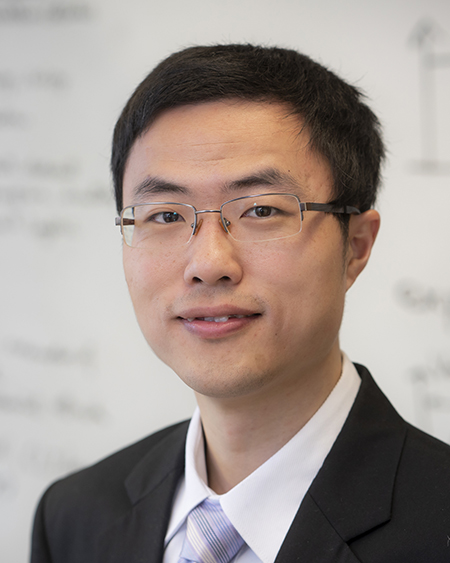}}]{Stanley H. Chan} (Senior Member, IEEE) is an Elmore Professor of Electrical and Computer Engineering at Purdue University, West Lafayette. His research interests include imaging through atmospheric turbulence, generative photography, and photon-limited imaging. Chan is a senior area editor of the IEEE Transactions on Computational Imaging. 
\end{IEEEbiography}

\vspace{-15mm}
\begin{IEEEbiography}
[{\includegraphics[width=1in,height=1.25in,clip,keepaspectratio]{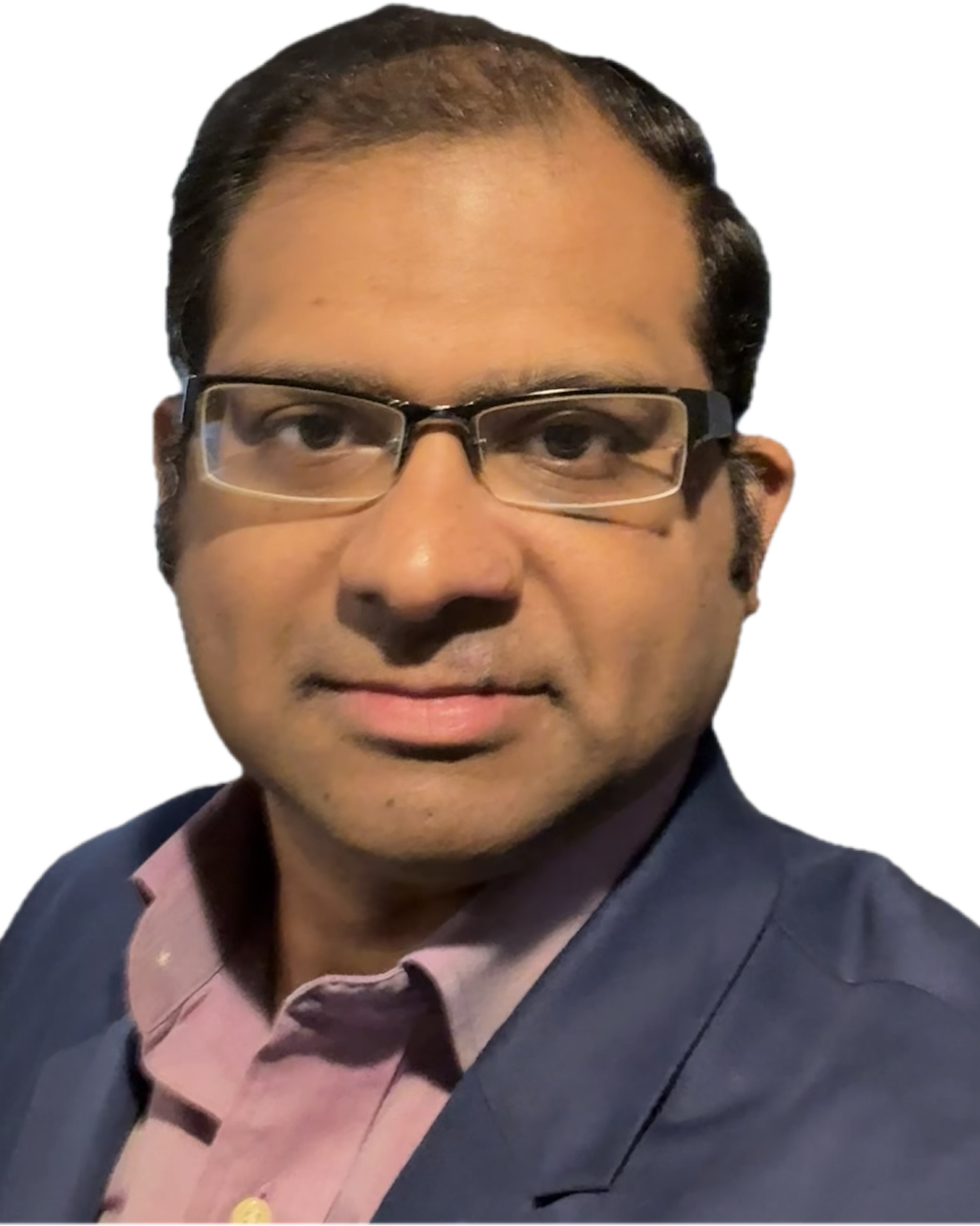}}]{Arun Ross} (Senior Member, IEEE) is the Martin J. Vanderploeg Endowed Professor in Computer Science and Engineering at Michigan State University and Site Director of NSF’s Center for Identification Technology Research (CITeR). Ross is the recipient of the NSF CAREER Award, the IAPR JK Aggarwal Prize, and the IAPR Young Biometrics Investigator Award. His research interests include biometrics, computer vision, and deep learning. 
%Ross is the recipient of the NSF CAREER Award, the IAPR JK Aggarwal Prize, and the IAPR Young Biometrics Investigator Award. He has advocated for the responsible use of biometrics in several forums including the NATO Advanced Research Workshop on Identity and Security in Switzerland in 2018. He testified as an expert panelist in an event organized by the United Nations Counter-Terrorism Committee at the UN Headquarters in 2013. In June 2022, he testified at the US House Science, Space, and Technology Committee on the topic of Biometrics and Personal Privacy. Ross is the co-author of ``Handbook of Multibiometrics" and the textbook ``Introduction to Biometrics".
\end{IEEEbiography}

\vspace{-15mm}
% if you ill not have a photo at all:
\begin{IEEEbiography}
[{\includegraphics[width=1in,height=1.25in,clip,keepaspectratio]{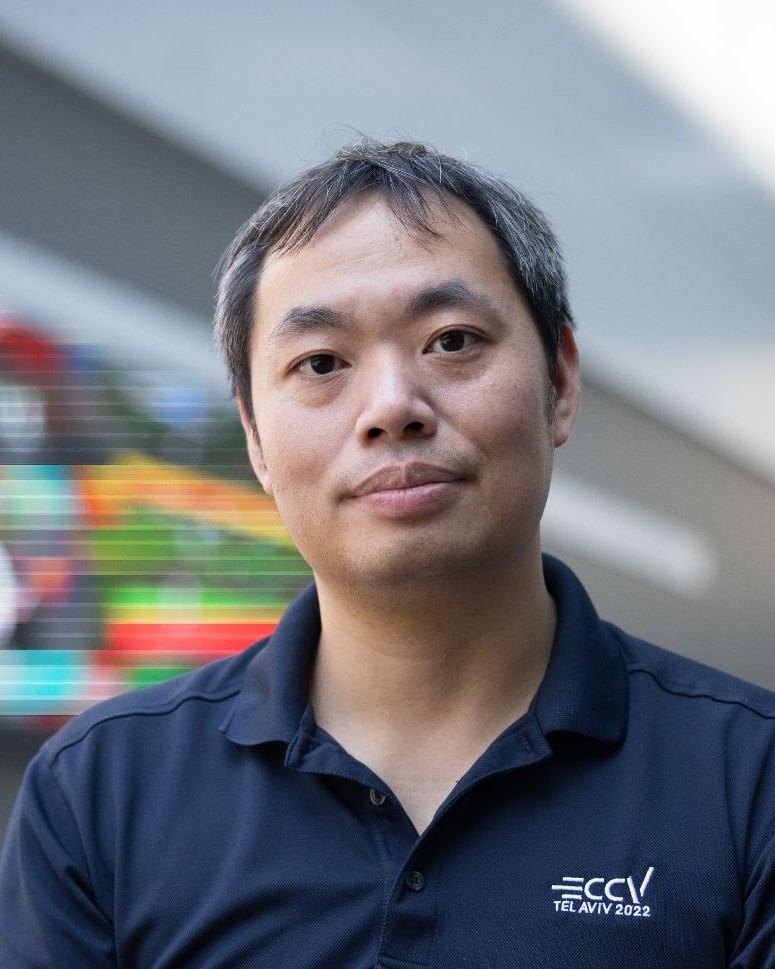}}]{Humphrey Shi} is an Associate Professor in the School of Interactive Computing and a member of the Machine Learning Center at Georgia Tech. 
% He is also a Graduate Faculty Member of Electrical and Computer Engineering at University of Illinois at Urbana-Champaign. Until recently, he served as the Chief Scientist of the popular image editing platform Picsart, through which he delivers advanced AI research and technologies to empower hundreds of millions of users globally. 
His current research focuses on building the next generation multimodal AI systems to understand, emulate, and interact with the world in a creative, efficient, and responsible way.
\end{IEEEbiography}

\vspace{-15mm}
\begin{IEEEbiography}
[{\includegraphics[width=1in,height=1.25in,clip,keepaspectratio]{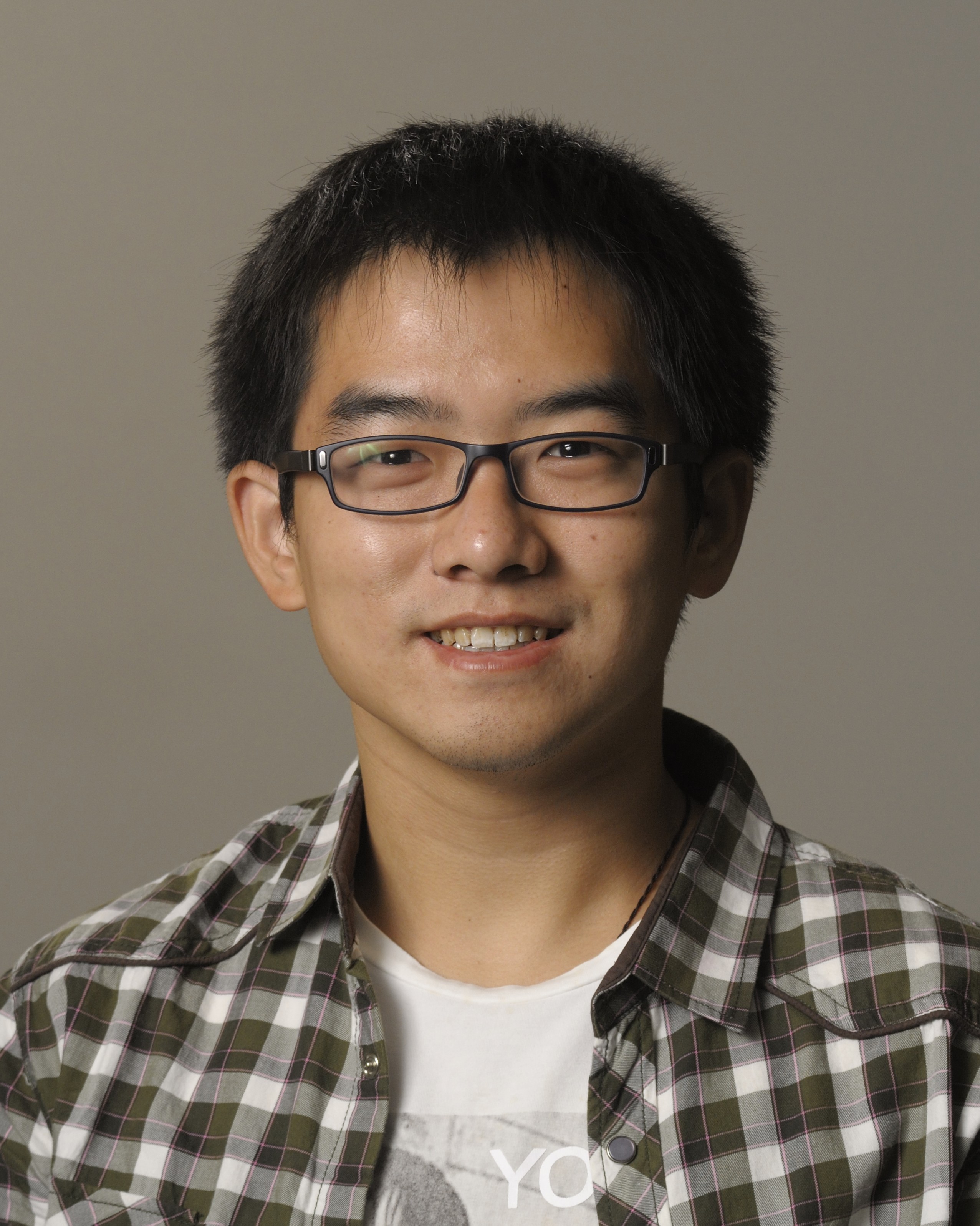}}]{Zhangyang Wang}
(\IEEEmembership{Senior Member,~IEEE}) is the Temple Foundation Endowed Associate Professor \#7 in the Chandra Family Department of Electrical and Computer Engineering at The University of Texas at Austin. His current research passion centers on developing the theoretical and algorithmic foundations of generative AI and neurosymbolic AI.
\end{IEEEbiography}

%\newpage
\vspace{-15mm}
\begin{IEEEbiography}
[{\includegraphics[width=1in,height=1.25in,clip,keepaspectratio]{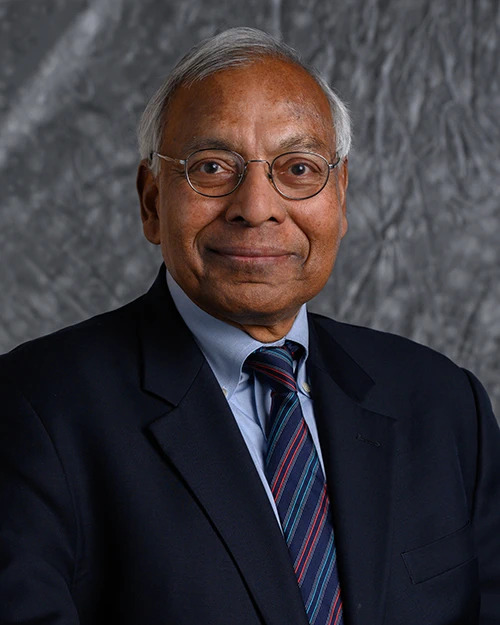}}]{Anil K. Jain}
(\IEEEmembership{Life Fellow,~IEEE}) is a University Distinguished Professor in the Department of Computer Science and Engineering at Michigan State University. His research interests include pattern recognition, computer vision, and biometric authentication. 
%He previously served as Editor-in-Chief of IEEE Transactions on Pattern Analysis and Machine Intelligence and was a member of the United States Defense Science Board. 
%He is the recipient of numerous prestigious honors, including the Fulbright Fellowship, Guggenheim Fellowship, Alexander von Humboldt Research Award, and the IAPR King-Sun Fu Prize.
%
Jain is a member of the U.S. National Academy of Engineering, the Indian National Academy of Engineering, the World Academy of Sciences, and the Chinese Academy of Sciences.
\end{IEEEbiography}

\vspace{-15mm}
\begin{IEEEbiography}
[{\includegraphics[width=1in,height=1.25in,clip,keepaspectratio]{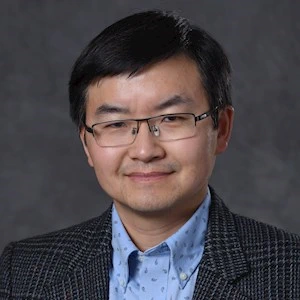}}]{Xiaoming Liu}
(\IEEEmembership{Fellow,~IEEE}) is a MSU Foundation Professor, and Anil and Nandita Jain Endowed Professor in the Department of Computer Science and Engineering at Michigan State University. He received his Ph.D.~from Carnegie Mellon University in 2004. His research interests span computer vision, machine learning, and biometrics.  
He is an Associate Editor for  IEEE Transactions on Pattern Analysis and Machine Intelligence. He is a fellow of IEEE and IAPR.
\end{IEEEbiography}


%\vfill
